\documentclass{article} 
\usepackage{iclr2025_conference,times}

\usepackage{amsmath,amsfonts,bm}

\def\mA{{\bm{A}}}
\def\mB{{\bm{B}}}
\def\mC{{\bm{C}}}
\def\mD{{\bm{D}}}
\def\mM{{\bm{M}}}
\newcommand{\R}{\mathbb{R}}

\usepackage{hyperref}
\usepackage{url}

\usepackage{booktabs}       
\usepackage{amsfonts}       
\usepackage{nicefrac}       
\usepackage{microtype}      
\usepackage{xcolor}         
\usepackage{natbib}         
\usepackage{makecell}       
\usepackage{multirow}       
\usepackage{multicol}
\usepackage{amsmath}        
\usepackage{graphicx}       
\usepackage{subcaption}  
\usepackage{color}
\usepackage[normalem]{ulem}
\usepackage{tikz}
\usepackage{colortbl}
\usepackage{xcolor}

\usepackage{xspace}
\makeatletter
\DeclareRobustCommand\onedot{\futurelet\@let@token\@onedot}
\def\@onedot{\ifx\@let@token.\else.\null\fi\xspace}

\def\ie{\emph{i.e}\onedot}

\usepackage{enumitem}
\setlist{itemsep=0pt}

\title{Spatial-Mamba: Effective Visual State Space\\ Models via Structure-aware State Fusion} 

\iclrfinalcopy
\author{Chaodong Xiao$^{1,2,^{\star}}$, Minghan Li$^{1,3,^{\star}}$, Zhengqiang Zhang$^{1,2}$, Deyu Meng$^{4}$, Lei Zhang$^{1,2,^{\dagger}}$\\
{$^{1}$The Hong Kong Polytechnic University \qquad $^{2}$OPPO Research Institute} \\
$^{3}$Harvard Medical School \hspace{2.75cm} $^{4}$Xi'an Jiaotong University\\
\texttt{chaodong.xiao@connect.polyu.hk, cslzhang@comp.polyu.edu.hk}\\
{\small $^{\star}$Equal contribution \qquad $^{\dagger}$Corresponding author}
}

\begin{document}

\maketitle
\vspace{-6pt}
\begin{abstract}

Selective state space models (SSMs), such as Mamba \citep{gu2023mamba}, highly excel at capturing long-range dependencies in 1D sequential data, while their applications to 2D vision tasks still face challenges. 
Current visual SSMs often convert images into 1D sequences and employ various scanning patterns to incorporate local spatial dependencies.
However, these methods are limited in effectively capturing the complex image spatial structures and the increased computational cost caused by the lengthened scanning paths.
To address these limitations, we propose \textbf{Spatial-Mamba}, a novel approach that establishes neighborhood connectivity directly in the state space. Instead of relying solely on sequential state transitions, we introduce a \textit{structure-aware state fusion} equation, which leverages dilated convolutions to capture image spatial structural dependencies, significantly enhancing the flow of visual contextual information.
Spatial-Mamba proceeds in three stages: initial state computation in a unidirectional scan, spatial context acquisition through structure-aware state fusion, and final state computation using the observation equation. Our theoretical analysis shows that Spatial-Mamba unifies the original Mamba and linear attention under the same matrix multiplication framework, providing a deeper understanding of our method.
Experimental results demonstrate that Spatial-Mamba, even with a single scan, attains or surpasses the state-of-the-art SSM-based models in image classification, detection and segmentation. 
Source codes and trained models can be found at \url{https://github.com/EdwardChasel/Spatial-Mamba}. 

\end{abstract}

\section{Introduction}

State space models (SSMs) are powerful tools for analyzing dynamic systems with hidden states, and they have long been utilized in fields like control theory, signal processing, and economics \citep{friston2003dynamic, hafner2019dream, gu2020hippo}. SSMs have been recently introduced into deep learning \citep{gu2021efficiently}, especially in natural language processing (NLP) \citep{gu2023mamba}, thanks to their use of specially parameterized matrices. A major advancement is the introduction of selective mechanisms and hardware-aware optimizations for parallel computing, as demonstrated by Mamba \citep{gu2023mamba}, which selectively retains or discards information based on the relevance of each element in a sequence, efficiently modeling long-distance dependencies with linear complexity.

\begin{figure}[t]
    \vspace{-10pt}
    \centering
    \begin{subfigure}{0.28\textwidth}
        \centering
        \includegraphics[height=3.0cm]{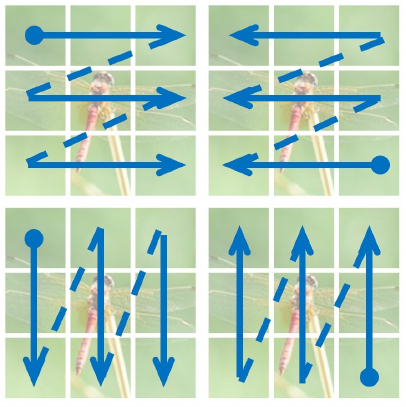}
        \caption{Sweeping Scan}
    \end{subfigure}
    \centering
    \begin{subfigure}{0.28\textwidth}
        \centering
        \includegraphics[height=3.0cm]{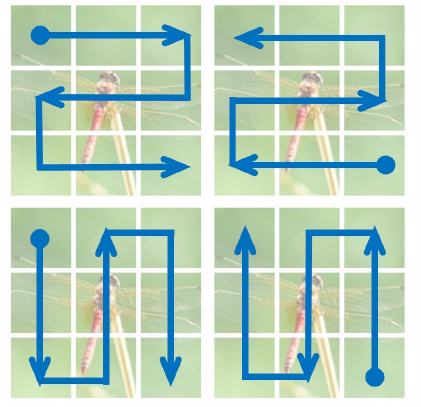}
        \caption{Continuous Scan}
    \end{subfigure}
    \centering
    \begin{subfigure}{0.28\textwidth}
        \centering
        \includegraphics[height=3.0cm]{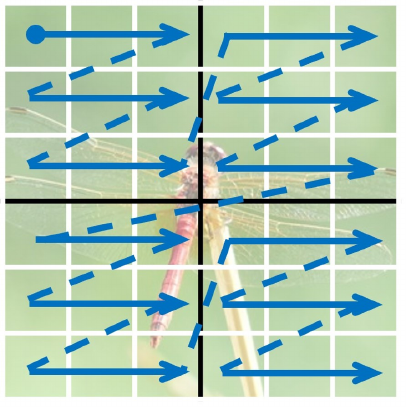}
        \caption{Local Scan}
    \end{subfigure}
    \vspace{0.01\textwidth}
        \\
     \centering
    \begin{subfigure}{0.9\textwidth}
        \centering
        \includegraphics[height=3.0cm]{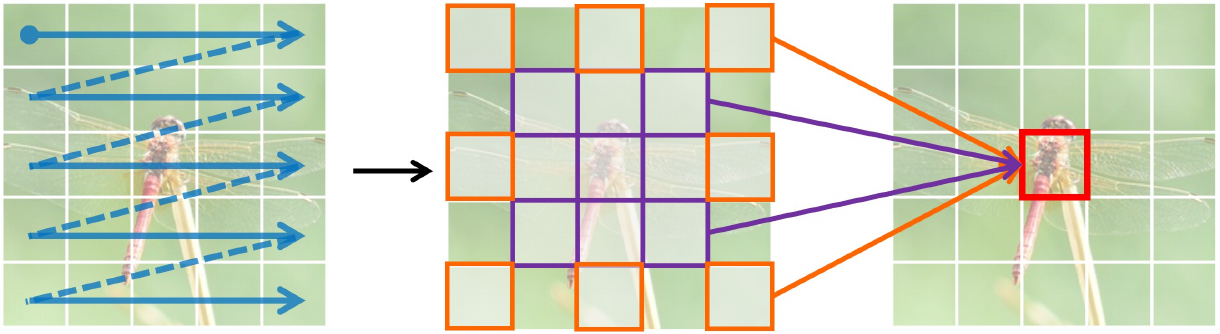}
        \caption{Structure-aware state fusion with unidirectional scan }
    \end{subfigure}

    \caption{Illustration of the scanning patterns of existing visual SSMs (from sub-figures (a) to (c)) and our proposed Spatial-Mamba with structure-aware state fusion (sub-figure (d)).}
    \vspace{-10pt}
    \label{fig:state}
\end{figure}

The significant success of Mamba in NLP inspires researchers to investigate how SSMs can be applied to visual tasks. Unlike 1D sequences, visual data are typically characterized by 2D spatial structures. 
Therefore, it is crucial to maintain the spatial dependencies within images while adapting the  information selection and propagation mechanisms in Mamba.
Existing visual SSMs \citep{zhu2024vision, liu2024vmamba, yang2024plainmamba,huang2024localmamba,  he2024mambaad, xiao2024grootvl} often use some scanning strategies to flatten 2D visual data into several 1D sequences from different directions, and  then process the flattened 1D sequences using the original Mamba. 
These scanning strategies can be broadly categorized into three types: sweeping scan, continuous scan and local scan, as shown in Figs.~\ref{fig:state}(a)-\ref{fig:state}(c), respectively.

Vim \citep{zhu2024vision} and VMamba \citep{liu2024vmamba} employ sweeping bidirectional scan and four-way scan, which are illustrated in Fig.~\ref{fig:state}(a). These strategies aim to reduce spatial direction sensitivity and adapt the network architecture for visual tasks.
\cite{yang2024plainmamba} argued that sweeping scans neglect the importance of spatial continuity, and they introduced a continuous scanning order, as shown in Fig.~\ref{fig:state}(b), to better integrate the inductive biases from visual data. 
\cite{huang2024localmamba} argued that scanning the entire image may not effectively capture local spatial relationships. Instead, they presented LocalMamba by using several local scanning modes, as shown in Fig.~\ref{fig:state}(c), which divides an image into distinct windows to capture local dependencies. 
Beyond the above mentioned scanning patterns, other scanning methods such as Hilbert scanning \citep{he2024mambaad} and dynamic tree scanning \citep{xiao2024grootvl} have also been proposed to adapt SSMs to visual tasks.

While the existing scanning strategies partially address the issue of aligning spatial structures with sequential SSMs, they are limited in model effectiveness and efficiency.
On one hand, directional scanning inevitably alters the spatial relationships between pixels, disrupting the inherent spatial context in an image. For example, in the sweeping scan (Fig.~\ref{fig:state}(a)), the distance between a pixel and its left or right neighbor is 1, while its distance to the top or bottom neighbor equals to the image width. This distortion can hinder the models to understand the spatial relationships in the original visual data. 
On the other hand, the fixed scanning paths, such as the commonly used four-directional scan (Figs.~\ref{fig:state}(a) and ~\ref{fig:state}(b)), are not effective enough to capture the complex and varying spatial relationships in an image, while introducing more scanning directions would also result in excessive computations. 
Therefore, it is imperative to explore how to design more effective and structure-aware SSMs for visual tasks.

To achieve this goal, we make an important attempt in this paper and present \textbf{Spatial-Mamba}, which is designed to capture the spatial dependencies of neighboring features in the latent state space. The processing flow of Spatial-Mamba consists of three stages. 
First, as shown in the left of Fig.~\ref{fig:state}(d), visual data are converted into sequential data using unidirectional sweeping scan. The state variables are computed based on the state transition equations of the original SSMs and then reshaped back into the visual format. 
Second, the state variables are processed through a \textbf{structure-aware state fusion} (SASF) equation, which employs dilated convolutions to re-weight and merge nearby state variables, as shown in the left of Fig.~\ref{fig:state}(d). 
Finally, these structure-aware state variables are fed into the observation equation to produce the final output variables. The SASF equation not only enables efficient skip connections between non-sequential elements in sequences but also enhances the model ability to capture spatial relationships, leading to more accurate representations of the underlying visual structure. 
Furthermore, we show that Spatial-Mamba, original Mamba and linear attention can all be represented under the same framework using structured  matrices, which offers a more coherent understanding of our proposed method. 
We validate the superiority of Spatial-Mamba across fundamental vision tasks such as image classification, detection, and segmentation. The results demonstrate that Spatial-Mamba, even with a single scan, achieves or surpasses the performance of recent state-of-the-arts using different scanning strategies.

\section{Related Work}

\textbf{State space models (SSMs).} 
\cite{gu2021combining} firstly introduced the linear state space layer (LSSL) into the HiPPO framework \citep{gu2020hippo} to efficiently handle the long-range dependencies in long sequences. 
\cite{gu2021efficiently} then significantly improved the efficiency of SSMs by representing the parameters as diagonal plus low-rank matrix. The so-called S4 model triggers a wave of structured SSMs \citep{smith2022simplified, fu2022hungry, gupta2022diagonal, gu2023mamba}.
\cite{smith2022simplified} proposed S5 by introducing parallel scans to S4 layer while maintaining the computational efficiency of S4.
Recently, \cite{gu2023mamba} developed Mamba, which incorporates a data-dependent selection mechanism into S4 layer and simplifies the computation and architecture in a hardware-friendly way, achieving Transformer-like modeling capability with linear complexity.
Building on that, \cite{dao2024transformers} presented Mamba2, which reveals the connections between SSMs and attention with specific structured matrix. This framework simplifies the parameter matrix to a scalar representation, making it feasible to explore larger and more expressive state spaces without sacrificing efficiency.

\textbf{Visual SSMs.}
Although traditional SSMs perform well in processing NLP sequential data and capturing temporal dependencies, they struggle in handling multi-dimensional spatial structure inherent in visual data. This limitation poses a challenge for developing effective visual SSMs. S4ND \citep{nguyen2022s4nd} is among the first SSM-based models for multi-dimensional data, which separates each dimension with an independent 1D SSM. \citet{baron20232} generalized S4ND as a discrete multi-axial system and proposed the 2D-SSM spatial layer, successfully extending the 1D SSMs to 2D SSMs. More recent visual SSMs prefer to design multiple scanning orders or patterns to maintain the spatial consistency, including bidirectional \citep{liu2024vmamba}, four-way \citep{zhu2024vision}, continuous \citep{yang2024plainmamba}, zigzag \citep{hu2024zigma}, window-based \citep{huang2024localmamba}, and topology-based scanning \citep{he2024mambaad,xiao2024grootvl}. 
These visual SSMs have been used in multimodal foundation models \citep{qiao2024vl,mo2024scaling}, image restoration \citep{guo2024mambair,shi2024vmambair}, medical image analysis \citep{yue2024medmamba,ma2024umamba,he2024densemamba,liao2024lightm} and other visual tasks \citep{chen2024video,li2024videomamba,yao2024spectralmamba}, demonstrating the potential of SSMs in visual data understanding.

\section{Preliminaries}
SSMs are commonly used for analyzing sequential data and modeling continuous linear time-invariant (LTI) systems \citep{2007Linear}. An input sequence $\displaystyle u(t) \in \R$ is transformed into an output sequence $\displaystyle y(t)\in \R$ through a state variable $\displaystyle x(t) \in \mathbb{C}^N$. Here, $t>0$ represents the time index, and $N$ indicates the dimension of the state variable. This dynamic system can be described by the linear state transition and observation equations \citep{kalman1960new}:
    $\displaystyle x'(t)  = \mA x(t) +  \mB u(t),y(t)  = \mC x(t) + \mD u(t)$,
where $\displaystyle \mA\in \mathbb{C}^{N\times N}$ is the state transition matrix, $\displaystyle \mB$, $\displaystyle \mC \in \mathbb{C}^{N}$ and $\displaystyle \mD \in \mathbb{C}^{1}$ control the dynamics of the system. The state transition and observation equations describe how the system evolves over time and how the state variables relate to the observed outputs.

To effectively integrate continuous-time SSMs into the deep learning framework, it is essential to discretize the continuous-time models. One commonly employed technique is the Zero-Order Hold (ZOH) discretization \citep{gu2023mamba}. The ZOH method approximates the continuous-time system by holding the input constant over each discrete time interval. Specifically, given a timescale $\mathbf{\Delta}$, which represents the interval between discrete time steps, and defining $\overline{\mA}$ and $\overline{\mB}$ as discrete parameters, the ZOH rule is applied as $\displaystyle \overline{\mA} =e^{\mathbf{\Delta} \mA}$ and $\overline{\mB}=(\mathbf{\Delta}\mA)^{-1}(e^{\mathbf{\Delta} \mA}-I)\mathbf{\Delta} \mB$.

However, real-world processes often change over time and cannot be accurately described by a LTI system. As highlighted in Mamba \citep{gu2023mamba}, time-varying systems can focus more on relevant information and offer a more accurate and realistic representation of dynamic systems. 
In Mamba, the parameters of the SSMs are made context-aware and adaptive through selective functions. This is achieved by modifying the  parameters $\mathbf{\Delta}, \mB,\mC $ as simple functions of the input sequence $u_t$, resulting in input-dependent parameters $\mathbf{\Delta}_t=s_\Delta(u_t), \mB_t=s_B(u_t)$ and $\mC_t=s_C(u_t)$. Then the input-dependent discrete parameters $\overline{\mA}_t$ and $\overline{\mB}_t$ can be calculated accordingly. Consequently, the discrete state transition and observation equations can be calculated as follows:
\begin{equation}\label{eq:ssm}
    \displaystyle
    x_t  = \overline{\mA}_t x_{t-1} + \overline{\mB}_t u_t, \quad
    y_t  = \mC_t x_t + \mD u_t.
\end{equation}

\begin{figure}[t]
    \centering
    \begin{subfigure}{0.45\textwidth}
        \centering
        \includegraphics[height=3.1cm]{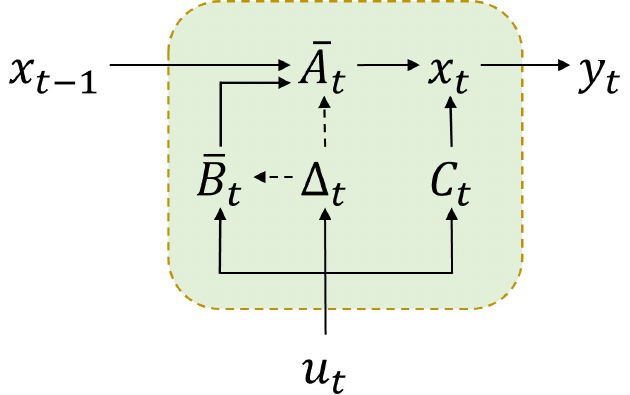}
        \caption{SSM in Mamba}
    \end{subfigure}
    \hfill
    \begin{subfigure}{0.45\textwidth}
        \centering
        \includegraphics[height=3.1cm]{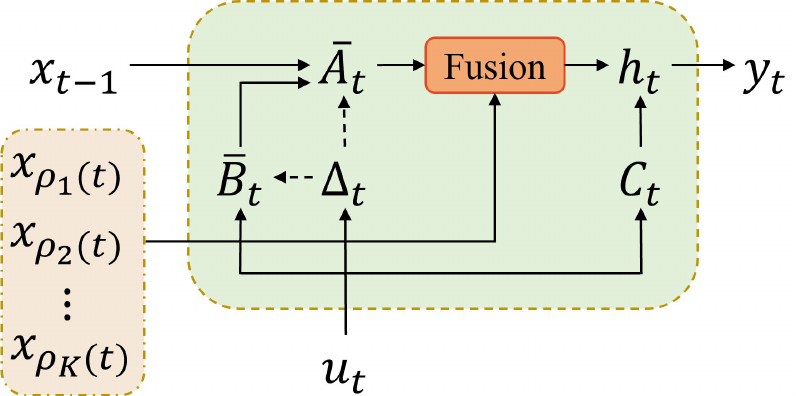}
        \caption{SSM in Spatial-Mamba}
    \end{subfigure}
    \caption{Illustrations of the SSM in (a) Mamba and (b) our Spatial-Mamba, where the residual term $\displaystyle \mD$ is omitted. In (b), `Fusion' refers to our proposed structure-aware state fusion (SASF) equation.}
    \label{fig:framework}
\end{figure}

A simplified illustration of the above process is shown in Fig.~\ref{fig:framework}(a).

\begin{figure}[t]
    \centering
    \begin{subfigure}{0.19\textwidth}
        \centering
        \includegraphics[height=2.2cm]{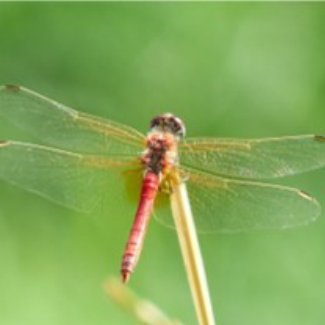}
        \caption{Input}
    \end{subfigure}
    \hfill
    \begin{subfigure}{0.19\textwidth}
        \centering
        \includegraphics[height=2.2cm]{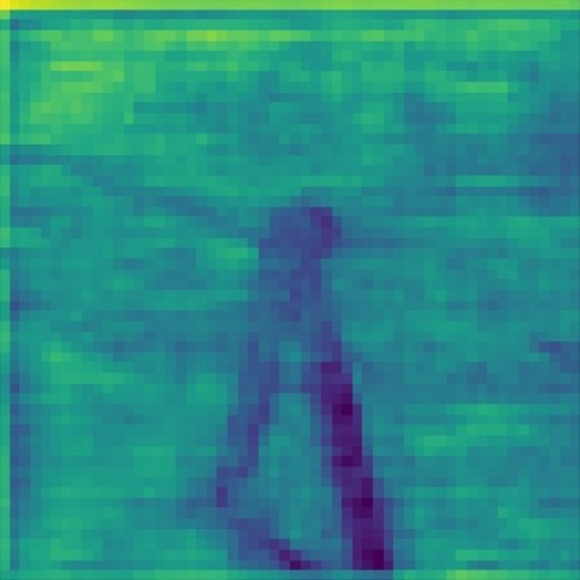}
        \caption{Original state $x_t$}
        \label{fig:original state}
    \end{subfigure}
    \hfill
    \begin{subfigure}{0.19\textwidth}
        \centering
        \includegraphics[height=2.2cm]{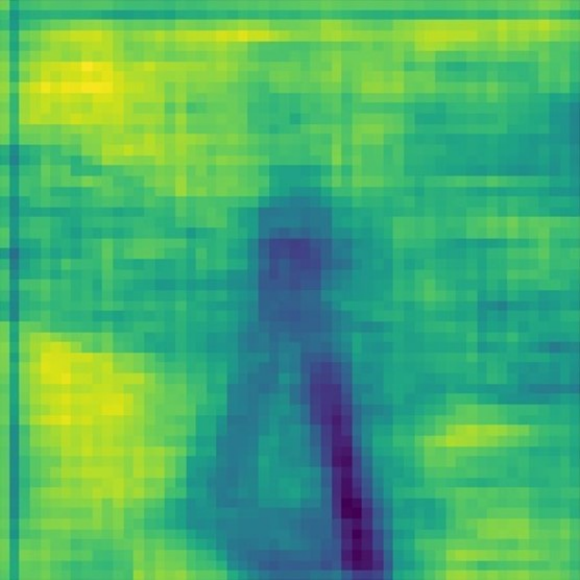}
        \caption{Fused state $h_t$}
        \label{fig:structure-aware state}
    \end{subfigure}
    \hfill
    \begin{subfigure}{0.01\textwidth}
        \centering
        \includegraphics[height=2.7cm]{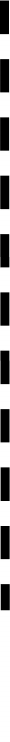}
    \end{subfigure}
    \hfill
    \begin{subfigure}{0.19\textwidth}
        \centering
        \includegraphics[height=2.2cm]{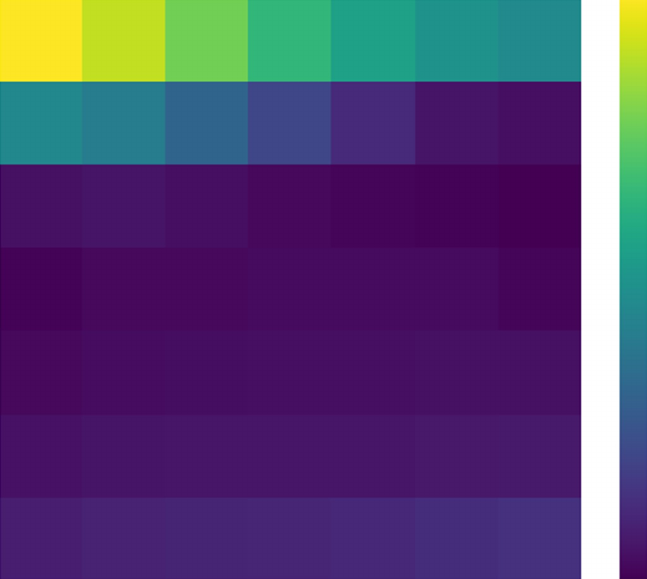}
        \caption{Original state $x_t$}
        \label{fig:original state 1}
    \end{subfigure}
    \hfill
    \begin{subfigure}{0.19\textwidth}
        \centering
        \includegraphics[height=2.2cm]{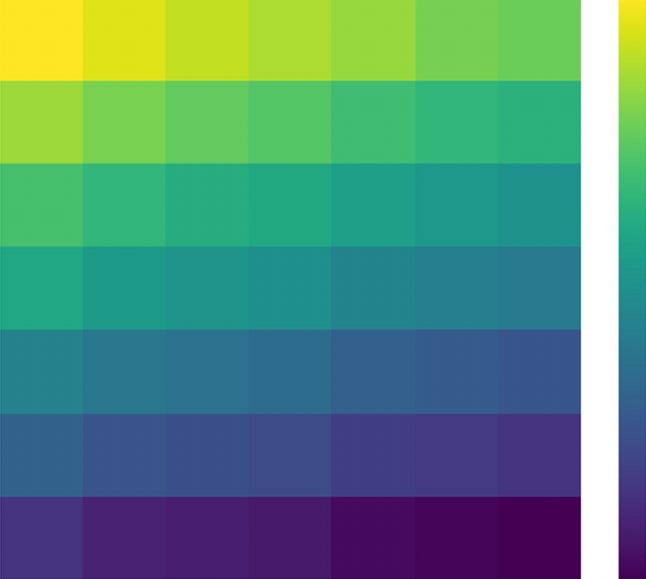}
        \caption{Fused state $h_t$}
        \label{fig:structure-aware state 1}
    \end{subfigure}
    \caption{Visualization of state variables before and after applying the SASF equation. Sub-figures (b) and (c) show the mean of state variables across all channels in the first layer of Spatial-Mamba, while sub-figures (d) and (e) display the state variables for a specific channel in the last layer.}
    \label{fig:statevs}
\end{figure}

\section{Spatial-Mamba}
\subsection{Formulation of Spatial-Mamba} \label{sec4.1}

Spatial-Mamba is designed to capture the spatial dependencies of neighboring features in the latent state space. To achieve this goal, different from previous methods \citep{zhu2024vision, liu2024vmamba,huang2024localmamba} that commonly employ multiple scanning directions, we introduce a new structure-aware state fusion (SASF) equation into the original Mamba formulas (refer to Eq.~(\ref{eq:ssm})). The entire process of Spatial-Mamba can be described by three equations: the state transition equation, the SASF and the observation equation, which are formulated as:
\begin{equation}\label{eq:state_fusion}
    \displaystyle
    x_t  = \overline{\mA}_t x_{t-1} + \overline{\mB}_t u_t, \quad
    h_t  = \sum_{k\in \Omega}\alpha_k x_{\rho_{k}(t)},  \quad
    y_t  = \mC_t h_t + \mD u_t,
\end{equation} 
where $x_t$ is the original state variable, $h_t$ is the structure-aware state variable, $\Omega$ is the neighbor set, $\alpha_k$ is a learnable weight, and $\rho_{k}(t)$ is the index of the $k$-th neighbor of position $t$. Fig.~\ref{fig:framework}(b) illustrates the SSM flow in the proposed Spatial-Mamba. Compared with the original Mamba in Fig.~\ref{fig:framework}(a), we can see that the original state variable $x_t$ is directly influenced by its previous state $x_{t-1}$, while the structure-aware state variable $h_t$ incorporates additional neighboring state variables $x_{\rho_1(t)},x_{\rho_2(t)},\dots,x_{\rho_K(t)}$ through a fusion mechanism, where $K=|\Omega|$ denotes the size of neighbor set. 
By considering both the global long-range and the local spatial information,
the fused state variable $h_t$ gains a richer context, leading to improved adaptability and a more comprehensive understanding of the image.

The proposed Spatial-Mamba can be implemented in three steps. As shown in Fig.~\ref{fig:state}(d), the input image is first flattened into a 1D sequence $u_t$, with which the state $x_t$ is computed using the state transition equation $ x_t  = \overline{\mA}_t x_{t-1} + \overline{\mB}_t u_t$. The computed states are then reshaped back into the 2D format.
To enable each state to be aware of its neighboring states in the 2D space, we introduce the SASF equation $h_t  = \sum_{k\in \Omega}\alpha_k x_{\rho_{k}(t)}$. For a state variable $ x_t $, we  apply linear weighting to its neighboring states $ \rho_{k}(t) $ in the neighborhood $ \Omega $ using weights $ \alpha_k $ so that we can effectively integrate local dependency information into a new state $ h_t $, resulting in a more contextually informative representation. Finally, the output is generated from this enriched state $h_t$ through the observation equation $y_t  = \mC_t h_t + \mD u_t$. This SASF approach helps the model to incorporate the local structural information in visual learning while retaining the benefits of original Mamba.

In practice, we simply employ multi-scale dilated convolutions \citep{yu2015multi} to linearly weight adjacent state variables, enhancing spatial relationship characterizations and enabling skip connections. Specifically, we use three $3 \times 3$ depth-wise filters with dilation factors $d = 1, 3, 5$ to construct the neighbor set $\Omega_d=\{(i,j)|i,j \in \{-d,0,d\}\}$. The SASF equation in Eq.~(\ref{eq:state_fusion}) can be rewritten as:
\begin{equation}
    h_t = \sum_{d=1,3,5 } \sum_{i,j\in \Omega_d} k_{ij}^d \cdot x_{t+iw+j},
\end{equation}
where $k_{ij}^d$ represents the filter weight for dilation factor $d$ at position $(i, j)$, $x_{t+iw+j}$ denotes the neighbor of the state $x_t$ located at the position $(i, j)$, and $w$ indicates the width of the image.

To gain an intuitive understanding of SASF, we visualize the original state variables and our proposed structure-aware state variables in Fig.~\ref{fig:statevs}. Additional visualizations are provided in Appendix~\ref{sec:appendix:morevisual}. One can see that the original state $x_t$ (Fig.~\ref{fig:statevs}(b)) struggles to differentiate the foreground from background. In contrast, the structure-aware state $h_t$ (Fig.~\ref{fig:statevs}(c)), which has been refined through a SASF process, effectively separates these regions. Moreover, the original state variable $x_t$ in Fig.~\ref{fig:statevs}(d) only shows horizontal attenuation along the scanning direction (gradually darkening from the brightest value in the upper left corner), while the fused state variable $h_t$ in Fig.~\ref{fig:statevs}(e) demonstrates decay along the horizontal, vertical and diagonal directions. This improvement stems from its ability to leverage spatial relationships within the image, leading to a more accurate and context-aware representation.

\subsection{Network Architecture}

\begin{figure}[t]
    \centering
    \includegraphics[width=\textwidth]{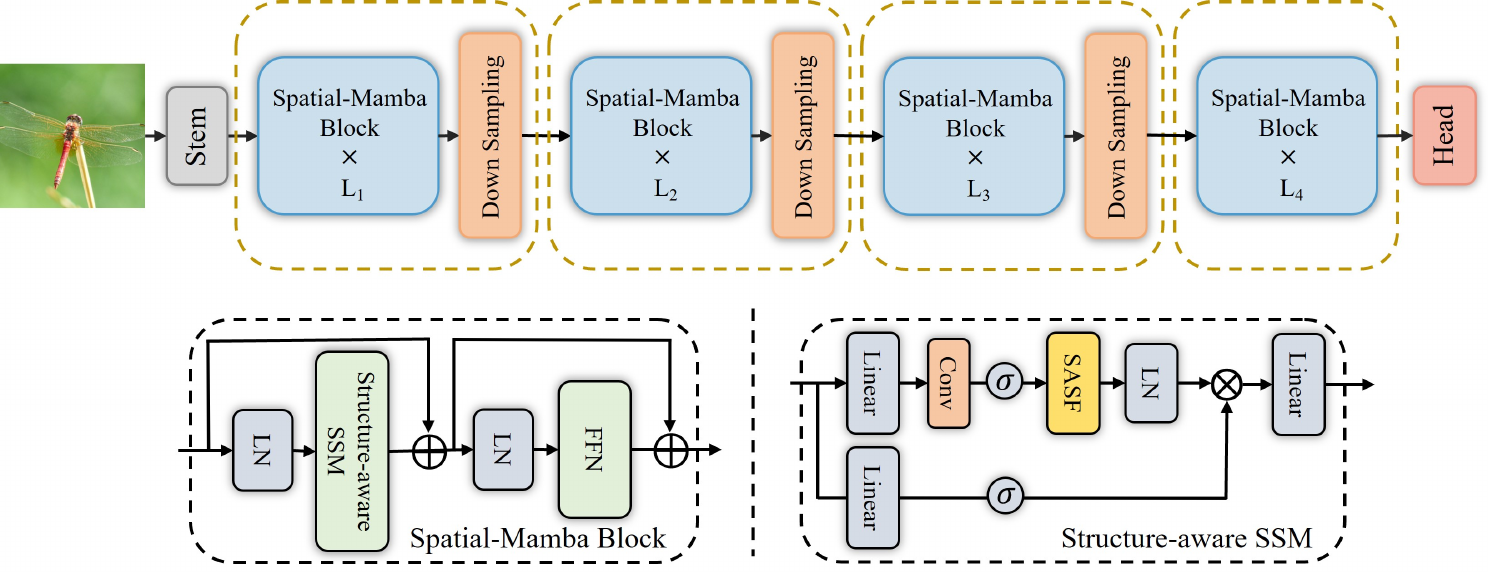}
    \caption{Overall network architecture of Spatial-Mamba. }
    \label{fig:architecture}
\end{figure}

The overall architecture of Spatial-Mamba is depicted in Fig.~\ref{fig:architecture}. It consists of four successive stages, resembling the structure of Swin-Transformer \citep{liu2021swin}. We introduce three variants of Spatial-Mamba model at different scales: Spatial-Mamba-T (tiny), Spatial-Mamba-S (small), and Spatial-Mamba-B (base). The detailed configurations are provided in Appendix~\ref{sec:appendix:detailed}.  
Specifically, an input image $I \in \mathbb{R}^{H\times W \times 3}$ is first processed by an overlapped stem layer to generate a 2D feature map with dimension of $\frac{H}{4}\times \frac{W}{4}\times C$. This feature map is then fed into four successive stages. Each stage comprises multiple Spatial-Mamba blocks, followed by a down-sampling layer with a factor of 2 (except for the last stage), resulting in hierarchical features. Finally, a head layer is employed to process these features to produce corresponding image representations for specific downstream tasks.

The Spatial-Mamba block forms the fundamental building unit of our architecture, which consists of a Structure-aware SSM and a Feed-Forward Network (FFN) with skip connections, as illustrated in the bottom left of Fig.~\ref{fig:architecture}. Building upon the Mamba block design \citep{gu2023mamba}, the Structure-aware SSM, illustrated in the bottom right of Fig.~\ref{fig:architecture}, is implemented by substituting the original 1D causal convolution with a $3\times3$ depth-wise convolution and replacing the original S6 module with our proposed SASF module, achieving local neighborhood connectivity in state spaces with linear complexity. Moreover, a local perception unit (LPU) \citep{guo2022cmt} is employed before the Spatial-Mamba block and FFN to extract local information inside image patches.

\subsection{Connection with Original Mamba and Linear Attention}

We analyze in-depth the similarities and disparities among linear attention \citep{katharopoulos2020transformers}, original Mamba \citep{gu2023mamba} and our Spatial-Mamba, providing a better understanding of the working mechanism of our proposed method. Detailed derivations are provided in Appendix~\ref{sec:appendix:derivation}.

\textbf{Linear attention} is an improved self-attention (SA) mechanism, reducing the computational complexity of SA to linear by using a kernel function $\phi$. For an input sequence $u_t$, the query $q_t$, key $k_t$, and value $v_t$ are computed by projecting $u_t$ with different weight matrices. In autoregressive models, to prevent the model from attending to future tokens, the $t$-th query is restricted by the previous keys, \ie, $k_s,s\leq t$. Thus, if the kernel function $\phi$ is an identity map, the  calculation of single head linear attention without normalization can be formulated as: $y_t=\sum_{s \leq t} q_t (k_s^T v_s)$. Letting $x_t=\sum_{s\leq t}k_s^T v_s$, then we have $x_t=x_{t-1}+k_t^T v_t$. The linear attention can be rewritten as $y_t=q_t x_t$. This reveals that linear attention is actually a special case of linear recursion. The hidden state variable $x_t$ is updated by the outer-product $k_s^T v_s$, and the final output $y_t$ is observed by multiplying $x_t$ with the query $q_t$. If we define $\mC_t=q_t$ and $\overline{\mB}_s= k_s^T$, the linear attention can be expressed in a form similar to that of SSM:
\begin{equation}\label{eq:linear attention}
    y_t=\sum_{s \leq t} \mC_t \overline{\mB}_s v_s,
\end{equation}
where $v_s$ is a linear transformation of $u_s$, \ie, $v_s=\text{Linear}(u_s).$

\textbf{Mamba} is essentially defined in Eq.~(\ref{eq:ssm}).
By setting the initial state variable $x_0$ to zero, the state variables can be derived recursively as $x_t = \sum_{s \leq t} \overline{\mA}_{s:t}^{\times} \overline{\mB}_s u_s$. Here, $\overline{\mA}_{s:t}^{\times}:=\Pi_{i=s+1}^t\overline{\mA}_{i}$ 
denotes the product of the state transition matrices with indices from $s+1$ to $t$ for $s<t$, and its value is $1$ when $s=t$. The final output of the observation equation without $\mD u_t$ can be rewritten as:
\begin{equation}\label{eq:ssm_mu}
    y_t = \sum_{s \leq t} \mC_t\ \overline{\mA}_{s:t}^{\times}\ \overline{\mB}_s u_s.
\end{equation}

\vspace{-4pt}
\textbf{Spatial-Mamba.} Based on Eq. (\ref{eq:state_fusion}) and Eq. (\ref{eq:ssm_mu}), the structure-aware state variables from our proposed SASF equation can be expressed as 
$h_t = \sum_{k \in \Omega}\ \sum_{s \leq \rho_k(t)} \alpha_k \overline{\mA}_{s:{\rho_k(t)}}^{\times}\ \overline{\mB}_s u_s $. By omitting $Du_t$ for simplicity of expression, the final output of Spatial-Mamba can be reformulated as follows: 
\begin{equation}
    y_t  = \sum_{k\in \Omega}\sum_{s \leq \rho_k(t)}\alpha_k\ \mC_t\ \overline{\mA}_{s:\rho_k(t)}^{\times} \ \overline{\mB}_s u_s.
\end{equation}

\begin{figure}[t]
    \vspace{-19pt}
    \centering
    \begin{subfigure}{0.13\textwidth}
        \centering
        \includegraphics[height=1.9cm]{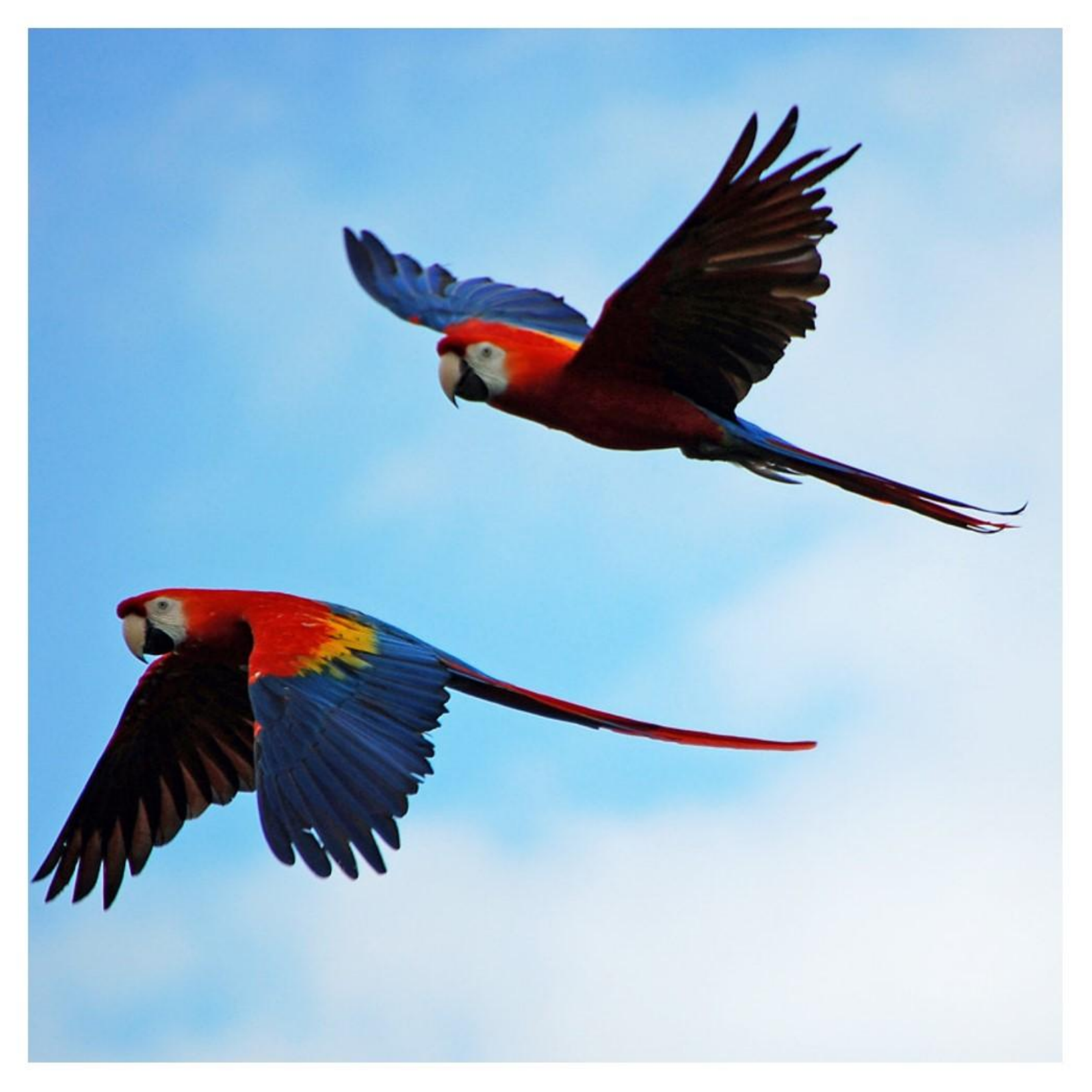}
        \caption{Input}
    \end{subfigure}
    \begin{subfigure}{0.28\textwidth}
        \centering
        \includegraphics[height=1.9cm]{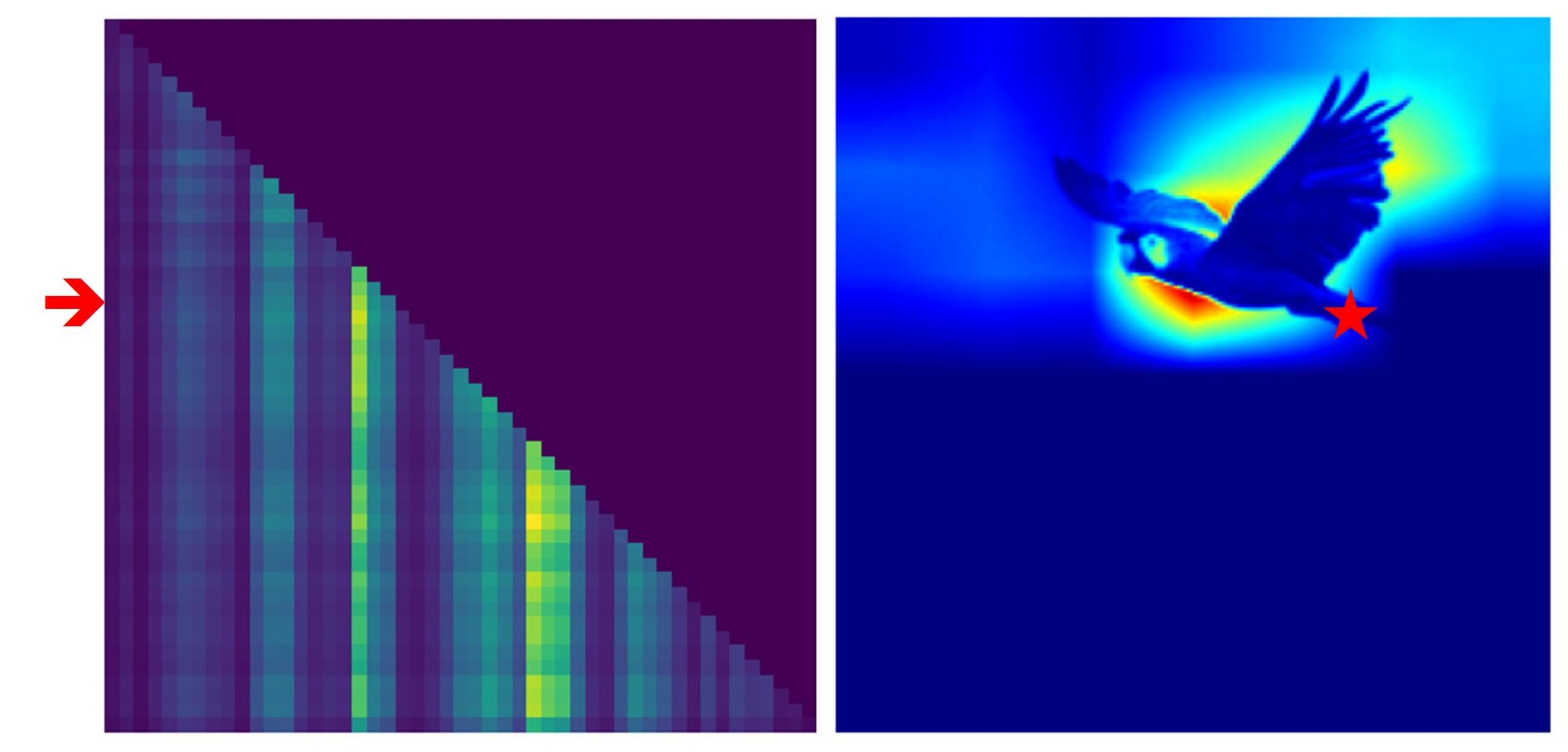}
        \caption{Linear attention}
    \end{subfigure}
    \begin{subfigure}{0.28\textwidth}
        \centering
        \includegraphics[height=1.9cm]{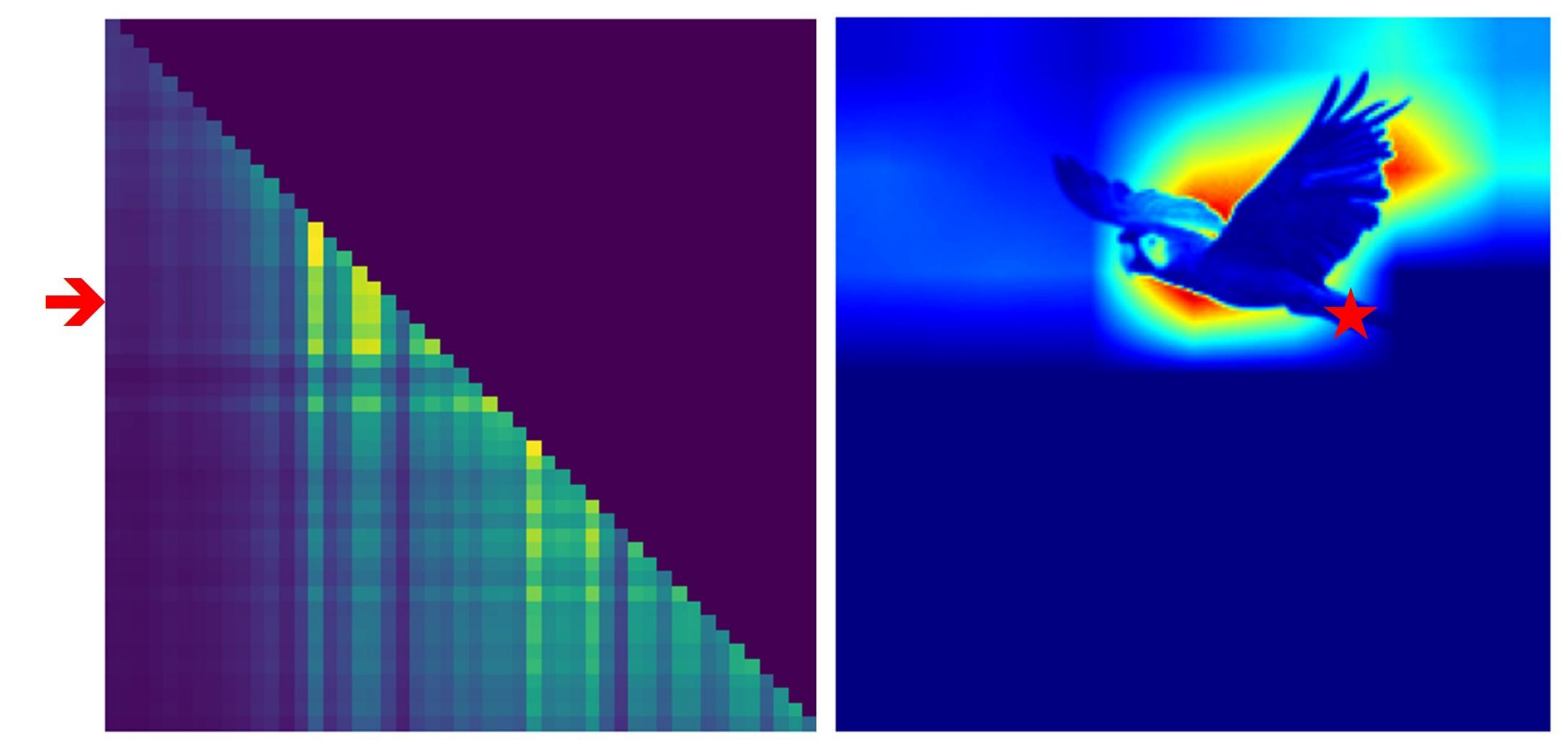}
        \caption{Mamba}
        \label{fig:M_Mamba}
    \end{subfigure}
    \begin{subfigure}{0.28\textwidth}
        \centering
        \includegraphics[height=1.9cm]{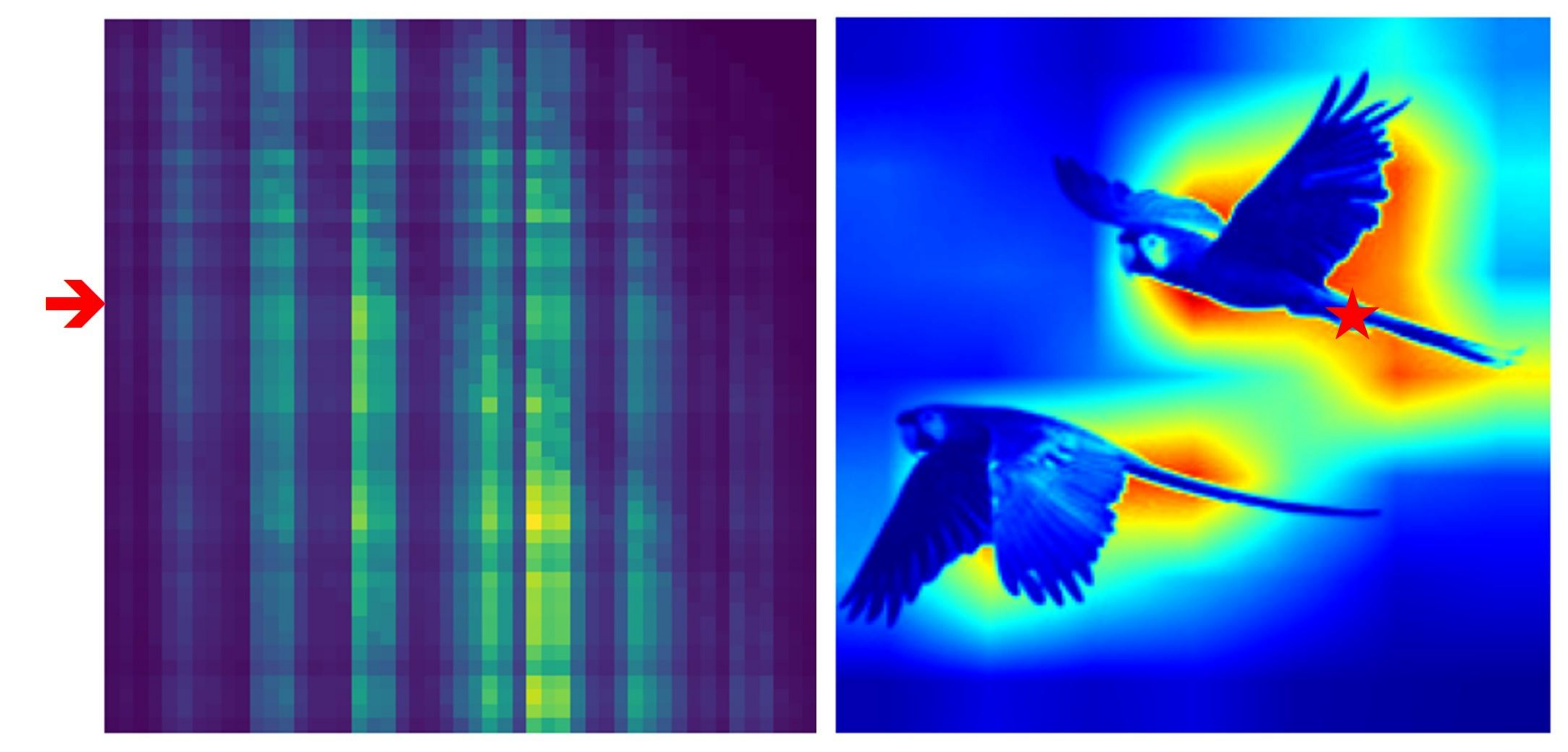}
        \caption{Spatial-Mamba}
        \label{fig:M-Spatial-Mamba}
    \end{subfigure}
    \caption{Visualizations of matrices $\mM$ and the corresponding activation maps for linear attention, Mamba and Spatial-Mamba. The red arrows indicate specific rows in matrices $\mM$, along with the corresponding image patches (marked with a red star).}
    \label{fig:matrixM}
    \vspace{-10pt}
\end{figure}
\vspace{-4pt}
\textbf{Remarks.} From the above analysis, we can see that all the three paradigms — linear attention, Mamba, and Spatial-Mamba — can be modeled within a unified matrix multiplication framework, specifically $y=\mM u$. The differences lie in the structure of $\mM$. For both linear attention and Mamba, $\mM$ takes the form of a \textbf{lower triangular matrix}, whereas for Spatial-Mamba, $\mM$ is an \textbf{adjacency matrix}. 
Fig. \ref{fig:matrixM} provides a visualization of these matrices. 
In linear attention, the positions of brighter values remain consistent along the vertical direction, which indicates that the SA mechanism puts its focus on a small set of image tokens. Mamba, on the other hand, shows a decaying pattern over time, which is attributed to the influence of its state transition matrix $\overline{\mA}_t$. This dynamic transition allows Mamba to shift its focus among previous image tokens. Unlike linear attention and Mamba, our Spatial-Mamba considers the weighted summation of all states within a broader spatial neighborhood $\Omega$, allowing for a more comprehensive representation of spatial relationships. The activation maps on the right side of Fig. \ref{fig:matrixM} further demonstrate that linear attention focuses on a limited region, while Mamba captures a broader region due to its long-range context modeling. Our Spatial-Mamba not only largely extends the range of context modeling but also enables spatial structure modeling, effectively identifying relevant regions even when they are distant from each other.

\section{Experimental Results}
In this section, we conduct a series of experiments to compare Spatial-Mamba with leading benchmark models, including those based on Convolutional Neural Networks (CNNs) \citep{radosavovic2020regnet,liu2022convnext,yu2024mambaout}, Vision Transformers \citep{dosovitskiy2020vit,liu2021swin,wang2022pvt,hassani2023neighborhood}, and recent visual SSMs  \citep{nguyen2022s4nd,zhu2024vision,liu2024vmamba,huang2024localmamba}. Following previous works \citep{liu2021swin, liu2024vmamba}, we train three variants of Spatial-Mamba, namely Spatial-Mamba-T, Spatial-Mamba-S and Spatial-Mamba-B. The detailed configurations are provided in Appendix~\ref{sec:appendix:detailed}. 
The performance evaluation is conducted on fundamental visual tasks, including image classification, object detection, and semantic segmentation. 

\subsection{Image Classification on ImageNet-1K}

\begin{table}[!b]
    \centering
    \caption{ Comparison of classification performance on ImageNet-1K, where `Throughput' is measured using an A100 GPU with an input resolution of $224\times 224$. 
    }
    \setlength{\tabcolsep}{1.2mm}
        \begin{tabular}{c|c|cccc|c}
        \toprule
        Arch. & Method & \makecell{Im. size} & \makecell{\#Param.  (M)} & \makecell{FLOPs (G)} & Throughput$\uparrow$ &  \makecell{Top-1 acc.$\uparrow$} \\
        \midrule
        \multirow{6}{*}{\rotatebox{90}{CNN}}
        & RegNetY-4G  & $224^2$ & 21M & 4.0G  & -  & 80.0 \\
        & RegNetY-8G  & $224^2$ & 39M & 8.0G  & -  & 81.7 \\
        & RegNetY-16G & $224^2$ & 84M & 16.0G & -  & 82.9 \\
        & ConvNeXt-T & $224^2$ & 29M & 4.5G  & 1189  & 82.1 \\
        & ConvNeXt-S & $224^2$ & 50M & 8.7G  & 682   & 83.1 \\
        & ConvNeXt-B & $224^2$ & 89M & 15.4G & 435   & 83.8 \\
        \midrule
        \multirow{8}{*}{\rotatebox{90}{Transformer\quad}}
        & ViT-B/16 & $384^2$ & 86M  & 55.4G  & -  & 77.9 \\
        & ViT-L/16 & $384^2$ & 307M & 190.7G & -  & 76.5 \\
        & DeiT-S & $224^2$ & 22M & 4.6G  & 1759  & 79.8 \\
        & DeiT-B & $224^2$ & 86M & 17.5G & 500   & 81.8 \\
        & DeiT-B & $384^2$ & 86M & 55.4G & 498   & 83.1 \\
        & Swin-T & $224^2$ & 28M & 4.5G  & 1244 & 81.3 \\
        & Swin-S & $224^2$ & 50M & 8.7G  & 718  & 83.0 \\
        & Swin-B & $224^2$ & 88M & 15.4G & 458  & 83.5 \\
        & NAT-T & $224^2$ & 28M & 4.3G  & - & 83.2 \\
        & NAT-S & $224^2$ & 51M & 7.8G  & -  & 83.0 \\
        & NAT-B & $224^2$ & 90M & 13.7G & -  & 84.3 \\
        \midrule
        \multirow{11}{*}{\rotatebox{90}{SSM}}
        & S4ND-ConvNeXt-T & $224^2$ & 30M & - & 683 & 82.2 \\
        & S4ND-ViT-B & $224^2$      & 89M & - & 397 & 80.4 \\
        & ViM-S & $224^2$           & 26M & - & 811 & 80.5 \\
        & VMamba-T & $224^2$ & 30M & 4.9G  & 1686  & 82.6 \\
        & VMamba-S & $224^2$ & 50M & 8.7G  & 877   & 83.6 \\
        & VMamba-B & $224^2$ & 89M & 15.4G & 646   & 83.9 \\
        & LocalVMamba-T & $224^2$ & 26M & 5.7G   & 394  & 82.7 \\
        & LocalVMamba-S & $224^2$ & 50M & 11.4G  & 227  & 83.7 \\
        \rowcolor{blue!5} & Spatial-Mamba-T & $224^2$ & 27M & 4.5G &  1438 & \textbf{83.5} \\
        \rowcolor{blue!5} & Spatial-Mamba-S & $224^2$ & 43M & 7.1G &  988 & \textbf{84.6} \\
        \rowcolor{blue!5} & Spatial-Mamba-B & $224^2$ & 96M & 15.8G & 665 & \textbf{85.3} \\
        \bottomrule
        \end{tabular}
    \label{exp:classification}
\end{table}

\textbf{Settings.} We first evaluate the representation learning capabilities of Spatial-Mamba in image classification on ImageNet-1K \citep{deng2009imagenet}. We  adopted the experimental configurations used in previous works \citep{liu2021swin, liu2024vmamba}, which are detailed in Appendix~\ref{sec:appendix:detailed}. We compare our method with state-of-the-art approaches, including RegNetY \citep{radosavovic2020regnet}, ConvNeXt \citep{liu2022convnext}, ViT \citep{dosovitskiy2020vit}, DeiT \citep{touvron2021deit}, Swin \citep{liu2021swin}, NAT \citep{hassani2023neighborhood}, S4ND \citep{nguyen2022s4nd}, Vim \citep{zhu2024vision}, VMamba \citep{liu2024vmamba}, and LocalVMamba \citep{huang2024localmamba}.

\begin{table}[!t]
    \vspace{-10pt}
    \centering
    \caption{ Comparison of object detection and instance segmentation performance on COCO  with Mask R-CNN \citep{he2017mask} detector. FLOPs are calculated with input resolution of $1280\times 800$.} \label{tab:detection}
    \setlength{\tabcolsep}{1.7mm}
        \begin{tabular}{c|ccc|ccc|cc}
        \toprule
        \multicolumn{9}{c}{\textbf{Mask R-CNN 1$\times$ schedule}}\\
        \midrule
        Backbone & AP$^\text{b}\uparrow$ & AP$^\text{b}_{50}$$\uparrow$ & AP$^\text{b}_{75}$$\uparrow$ & AP$^\text{m}\uparrow$ & AP$^\text{m}_{50}$$\uparrow$ & AP$^\text{m}_{75}$$\uparrow$ & \#Param. & FLOPs \\
        \midrule
        ResNet-50  & 38.2 & 58.8 & 41.4 & 34.7 & 55.7 & 37.2 & 44M & 260G \\
        Swin-T     & 42.7 & 65.2 & 46.8 & 39.3 & 62.2 & 42.2 & 48M & 267G \\
        ConvNeXt-T & 44.2 & 66.6 & 48.3 & 40.1 & 63.3 & 42.8 & 48M & 262G \\
        PVTv2-B2   & 45.3 & 66.1 & 49.6 & 41.2 & 64.2 & 44.4 & 45M & 309G \\
        ViT-Adapter-S & 44.7 & 65.8 & 48.3 & 39.9 & 62.5 & 42.8 & 48M & 403G \\
        MambaOut-T     & 45.1 & 67.3 & 49.6 & 41.0 & 64.1 & 44.1 & 43M & 262G \\
        VMamba-T    & 47.3 & 69.3 & 52.0 & 42.7 & 66.4 & 45.9 & 50M & 271G \\
        LocalVMamba-T & 46.7 & 68.7 & 50.8 & 42.2 & 65.7 & 45.5 & 45M & 291G \\
        \rowcolor{blue!5} Spatial-Mamba-T   & \textbf{47.6} & \textbf{69.6} & \textbf{52.3} & \textbf{42.9} & \textbf{66.5} & \textbf{46.2} & 46M & 261G \\
        \midrule
        ResNet-101 & 38.2 & 58.8 & 41.4 & 34.7 & 55.7 & 37.2 & 63M & 336G \\
        Swin-S     & 44.8 & 68.6 & 49.4 & 40.9 & 65.3 & 44.2 & 69M & 354G \\
        ConvNeXt-S & 45.4 & 67.9 & 50.0 & 41.8 & 65.2 & 45.1 & 70M & 348G \\
        PVTv2-B3   & 47.0 & 68.1 & 51.7 & 42.5 & 65.2 & 45.7 & 63M & 397G \\
        MambaOut-S      & 47.4 & 69.1 & 52.4 & 42.7 & 66.1 & 46.2 & 65M & 354G \\
        VMamba-S    & 48.7 & 70.0 & 53.4 & 43.7 & 67.3 & 47.0 & 70M & 349G \\
        LocalVMamba-S & 48.4 & 69.9 & 52.7 & 43.2 & 66.7 & 46.5 & 69M & 414G \\
        \rowcolor{blue!5} Spatial-Mamba-S   & \textbf{49.2} & \textbf{70.8} & \textbf{54.2} & \textbf{44.0} & \textbf{67.9} & \textbf{47.5} &63M & 315G \\
        \midrule
        Swin-B     & 46.9 & -    & -    & 42.3 & 66.3 & 46.0 & 88M & 496G \\
        ConvNeXt-B & 47.0 & 69.4 & 51.7 & 42.7 & 66.3 & 46.0 & 107M & 486G \\
        PVTv2-B5   & 47.4 & 68.6 & 51.9 & 42.5 & 65.7 & 46.0 & 102M & 557G \\
        ViT-Adapter-B & 47.0 & 68.2 & 51.4 & 41.8 & 65.1 & 44.9 & 102M & 557G \\
        MambaOut-B      & 47.4 & 69.3 & 52.2 & 43.0 & 66.4 & 46.3 & 100M & 495G  \\
        VMamba-B    & 49.2 & 71.4 & 54.0 & 44.1 & 68.3 & 47.7 & 108M & 485G 
        \\
        \rowcolor{blue!5} Spatial-Mamba-B   & \textbf{50.4} & \textbf{71.8} & \textbf{55.3} & \textbf{45.1} & \textbf{69.1} & \textbf{49.1} & 115M & 494G\\
        \toprule
        \multicolumn{9}{c}{\textbf{Mask R-CNN 3$\times$ MS schedule}}\\
        \midrule
        Backbone & AP$^\text{b}\uparrow$ & AP$^\text{b}_{50}$$\uparrow$ & AP$^\text{b}_{75}$$\uparrow$ & AP$^\text{m}\uparrow$ & AP$^\text{m}_{50}$$\uparrow$ & AP$^\text{m}_{75}$$\uparrow$ & \#Param. & FLOPs \\
        \midrule
        Swin-T & 46.0 & 68.1 & 50.3 & 41.6 & 65.1 & 44.9 & 48M & 267G \\
        ConvNeXt-T & 46.2 &67.9 &50.8& 41.7& 65.0& 44.9 &48M& 262G \\
        NAT-T & 47.7 &69.0& 52.6 &42.6& 66.1& 45.9& 48M & 258G \\
        VMamba-T &48.8 &70.4& 53.5& 43.7& 67.4& 47.0& 50M& 271G \\
        LocalVMamba-T &48.7 &70.1& 53.0& 43.4& 67.0& 46.4& 45M& 291G \\
        \rowcolor{blue!5} Spatial-Mamba-T   & \textbf{49.3} & \textbf{70.7} & \textbf{54.3} & \textbf{43.8} & \textbf{67.8} & \textbf{47.2} & 46M & 261G\\
        \midrule
        Swin-S &48.2 &69.8& 52.8& 43.2& 67.0& 46.1& 69M &354G \\
        ConvNeXt-S& 47.9& 70.0& 52.7& 42.9& 66.9& 46.2& 70M& 348G\\
        NAT-S  &48.4& 69.8 &53.2 &43.2 &66.9 &46.5 &70M &330G\\
        VMamba-S &49.9 &70.9& 54.7& 44.2& 68.2& 47.7 &70M &349G\\
        LocalVMamba-S & 49.9& 70.5& 54.4& 44.1& 67.8 &47.4 &69M &414G \\ 
        \rowcolor{blue!5} Spatial-Mamba-S & \textbf{50.5} & \textbf{71.5} & \textbf{55.5} & \textbf{44.6} & \textbf{68.7} & \textbf{47.8} & 63M & 315G\\
        \bottomrule 
        \end{tabular}
\end{table}

\textbf{Results.} Tab.~\ref{exp:classification} presents a comprehensive comparison between Spatial-Mamba against state-of-the-art methods. Notably, Spatial-Mamba-T achieves a top-1 accuracy of 83.5\%, outperforming the CNN-based ConvNext-T by 1.4\% with similar amount of parameters and FLOPs. Compared to Transformer-based methods, Spatial-Mamba-T exceeds Swin-T by 2.2\% and NAT-T by 0.3\%. In comparison with SSM-based methods, Spatial-Mamba-T outperforms VMamba-T by 1.0\% and LocalVMamba-T by 0.8\%.
For other variants, Spatial-Mamba also shows advantages. Specifically, Spatial-Mamba-S and Spatial-Mamba-B achieve top-1 accuracies of 84.6\% and 85.3\%, respectively, surpassing NAT-S and NAT-B by margins of 1.6\% and 1.0\%, and VMamba-S and VMamba-B by 1.0\% and 1.4\%. While Spatial-Mamba-T is slightly slower than VMamba-T due to architectural differences, the Small and Base variants of Spatial-Mamba are faster than their VMamba counterparts. Moreover, both of them are significantly faster than CNN and Transformer-based methods. 

\subsection{Object Detection and Instance Segmentation on COCO}
\textbf{Settings.} We evaluate Spatial-Mamba in object detection and instance segmentation tasks using COCO 2017 dataset \citep{lin2014microsoft} and MMDetection library \citep{chen2019mmdetection}. We adopt Mask \citep{he2017mask} and Cascade Mask R-CNN \citep{cai2018cascade} as detector heads, apply Spatial-Mamba-T/S/B pre-trained on ImageNet-1K as backbones. Following common practices \citep{liu2021swin, liu2024vmamba}, we fine-tune the pre-trained models for 12 epochs (1$\times$ schedule) and 36 epochs with multi-scale inputs (3$\times$ schedule). During training, AdamW optimizer is adopted with an initial learning rate of 0.0001 and a batch size of 16.

\textbf{Results.} 
The results on COCO with Mask R-CNN are reported in Tab.~\ref{tab:detection}, and the results with Cascade Mask R-CNN are provided in Appendix~\ref{sec:appendix:cascademask}. 
It can be seen that all variants of Spatial-Mamba outperform their competitors under different schedules. For 1$\times$ schedule, Spatial-Mamba-T achieves a box mAP of 47.6 and a mask mAP of 42.9, surpassing Swin-T/VMamba-T by 4.9/0.3 in box mAP and 3.6/0.2 in mask mAP with fewer parameters and FLOPs, respectively. Similarly, Spatial-Mamba-S/B demonstrate superior performance to other methods under the same configuration. Furthermore, these trends of improved performance hold with the 3$\times$ multi-scale training schedule. Notably, Spatial-Mamba-S achieves the highest box mAP of 50.5 and mask mAP of 44.6, surpassing VMamba-S with a considerable gain of 0.6 and 0.4, respectively.

\begin{table}[t]
    \vspace{-10pt}
    \centering
    \caption{Comparison of semantic segmentation on ADE20K with UPerNet \citep{xiao2018unified} segmentor. FLOPs are calculated with input resolution of $512 \times 2048$. `SS' and `MS' represent single-scale and multi-scale testing, respectively.}
    \label{tab:sem_seg}
    \setlength{\tabcolsep}{1.7mm}
        \begin{tabular}{c|c|cc|cc}
        \toprule
        Method & Crop size & mIoU (SS) $\uparrow$ & mIoU (MS) $\uparrow$ & \#Param. & FLOPs \\ 
        \midrule
        DeiT-S + MLN    & $512^2$ & 43.1 & 43.8 & 58M & 1217G  \\ 
        Swin-T          & $512^2$ & 44.4 & 45.8 & 60M & 945G   \\ 
        ConvNeXt-T      & $512^2$ & 46.0 & 46.7 & 60M & 939G   \\ 
        NAT-T           & $512^2$ & 47.1 & 48.4 & 58M &934G \\
        MambaOut-T        & $512^2$ & 47.4 & 48.6 & 54M & 938G  \\
        VMamba-T        & $512^2$ & 48.0 & 48.8 & 62M & 949G   \\ 
        LocalVMamba-T   & $512^2$ & 47.9 & 49.1 & 57M & 970G   \\ 
        \rowcolor{blue!5} Spatial-Mamba-T     & $512^2$ & \textbf{48.6} & \textbf{49.4} & 57M & 936G\\
        \midrule
        DeiT-B + MLN    & $512^2$ & 45.5 & 47.2 & 144M & 2007G  \\ 
        Swin-S          & $512^2$ & 47.6 & 49.5 & 81M  & 1039G  \\ 
        ConvNeXt-S      & $512^2$ & 48.7 & 49.6 & 82M  & 1027G  \\ 
        NAT-S           & $512^2$ & 48.0 & 49.5 & 82M & 1010G\\
        MambaOut-S        & $512^2$ & 49.5 & 50.6 & 76M  & 1032G \\
        VMamba-S        & $512^2$ & \textbf{50.6} & 51.2 & 82M  & 1028G  \\ 
        LocalVMamba-S   & $512^2$ & 50.0 & 51.0 & 81M  & 1095G  \\
        \rowcolor{blue!5} Spatial-Mamba-S & $512^2$ & \textbf{50.6} & \textbf{51.4} & 73M  & 992G\\
        \midrule
        Swin-B          & $512^2$ & 48.1 & 49.7 & 121M & 1188G  \\ 
        ConvNeXt-B      & $512^2$ & 49.1 & 49.9 & 122M & 1170G  \\ 
        NAT-B           & $512^2$ & 48.5 & 49.7 & 123M & 1137G \\
        MambaOut-B      & $512^2$ & 49.6 & 51.0 & 112M & 1178G  \\
        VMamba-B        & $512^2$ & 51.0 & 51.6 & 122M & 1170G  \\ 
        \rowcolor{blue!5} Spatial-Mamba-B & $512^2$ & \textbf{51.8} & \textbf{52.6} & 127M & 1176G \\
        \bottomrule
        \end{tabular}
\end{table}

\subsection{Semantic Segmentation on ADE20K}

\textbf{Settings.} To assess the performance of Spatial-Mamba on semantic segmentation task, we train our models with the widely used  UPerNet segmentor \citep{xiao2018unified} and MMSegmenation toolkit \citep{contributors2020mmsegmentation}. Consistent with previous work \citep{liu2021swin, liu2024vmamba}, we pre-train our model on ImageNet-1K, and use it as the backbone to train UPerNet on ADE20K dataset \citep{zhou2019semantic}. This training process encompasses 160K iterations with a batch size of 16. The AdamW is used as the optimizer with a weight decay of 0.01. The learning rate is set to $6\times 10^{-5}$ with a linear learning rate decay. All the input images are cropped into $512\times 512$.

\textbf{Results.} The results on semantic segmentation are summarized in Tab.~\ref{tab:sem_seg}. Spatial-Mamba variants consistently achieve impressive performance. For instance, Spatial-Mamba-T attains a single-scale mIoU of 48.6 and a multi-scale mIoU of 49.4. This signifies an improvement of 1.5 mIoU over NAT-T and 0.6 mIoU over VMamba-T in single-scale input. This advantage is maintained with multi-scale input, where Spatial-Mamba-T is 1.0 mIoU higher than NAT-T and 0.6 mIoU higher than VMamba-T. Furthermore, Spatial-Mamba-B achieves the best performance with a multi-scale mIoU of 52.6.

\subsection{Ablation Studies}
In this section, we ablate various key components of Spatial-Mamba-T on ImageNet-1K classification task in Tab.~\ref{table:ablationstudy}. Based on the configurations of Swin-T \citep{liu2021swin}, we construct the baseline model as Spatial-Mamba-T but without the SASF module. This baseline uses a $4\times4$ convolution with a stride of $4$ as stem layer and merges $2\times2$ neighboring patches for down-sampling.

\textbf{Neighbor set.} First, adjustments to the neighbor set $\Omega$ reveal that increasing the size from a $3\times3$ neighborhood to $5\times 5$ results in a gradual improvement in accuracy, from 82.3\% to 82.5\%, albeit with a corresponding decrease in throughput. Furthermore, employing a broader dilated neighbor set with factors $d=1,3,5$ increases the accuracy to 82.7\%, while reducing throughput to 1158. This suggests a trade-off between speed and larger neighbor set. 

\begin{table}[!t]
    \vspace{-10pt}
    \centering
    \caption{ Ablation studies on Spatial-Mamba-T for  neighbor set $\Omega$ and other module designs.} 
    \setlength{\tabcolsep}{1.7mm}
        \begin{tabular}{l|cccc}
        \toprule
        \ Module design  &\makecell{\#Param. (M)} & \makecell{FLOPs (G)} & Throughput $\uparrow$ &  \makecell{Top-1 acc.$\uparrow$} \\
        \midrule
        \ Baseline              &25M &  4.4G & 1706 & 82.0 \\ 
        \midrule
        \ $\Omega=\{3\times3\}$                    &25M &  4.4G & 1557 & 82.3 \\
        \ $\Omega=\{5\times5\}$                    &25M &  4.5G & 1461 & 82.5 \\
        \ $\Omega=\left\{\Omega_d|d=1,3,5\right\}$ &25M &  4.5G & 1209 & 82.7 \\
        \midrule
        
        \ $+$ Overlapped Stem  &27M &  4.5G & 1158 & 82.9 \\
        \ $+$ LPU              &27M &  4.5G & 1065 & 83.3 \\ 
        \ $+$ Re-Param         &27M &  4.5G & 1438 & 83.3 \\
        \ $+$ MESA             &27M &  4.5G & 1438 & 83.5 \\
        \bottomrule
        \end{tabular}
    \label{table:ablationstudy}
    \vspace{-5pt}
\end{table}

\textbf{Local enhancement.} We replace the original non-overlapped stem and down-sampling layer with overlapped convolutions (refer to as `Overlapped Stem' in Tab.~\ref{table:ablationstudy}), resulting in a gain of 0.2\% in accuracy. We also incorporate the LPU \citep{guo2022cmt}, a depth-wise convolution placed at the top of each block and FFN, further increasing the accuracy by 0.4\%. These modifications enrich the local information available between image patches before processing by the SASF module, enabling it to better capture structural dependencies.

\textbf{Optimization.} To further enhance the model efficiency, we implement the SASF module using re-parameterization techniques \citep{ding2022scaling} and optimize the CUDA kernels. This accelerates the model by at least 30\% and boosts the throughput from 1065 to 1438. Finally, integrating MESA \citep{du2022sharpness} to mitigate overfitting provides an additional 0.2\% accuracy improvement.

In addition, we provide some qualitative results in Appendix~\ref{sec:appendix:visual}, an in-depth discussion about SASF in Appendix~\ref{sec:appendix:extended ab study}, and a comparative analysis of the Effective Receptive Fields (ERF) \citep{ding2022scaling} of various models is provided in Appendix~\ref{sec:appendix:erf}.

\section{Conclusion}
We  presented in this paper Spatial-Mamba, a novel state space model designed for visual tasks. The key of Spatial-Mamba lied in the proposed structure-aware state fusion (SASF) module, which effectively captured image spatial dependencies and hence improved the contextual modeling capability. We performed extensive experiments on fundamental vision tasks of image classification, detection and segmentation. The results showed that with SASF, Spatial-Mamba surpassed the state-of-the-art state space models with only one signal scan, demonstrating its strong visual feature learning capability. We also analyzed in-depth the relationships of Spatial-Mamba with the original Mamba and linear attention, and unified them under the same matrix multiplication framework, offering a deeper understanding of the self-attention mechanism for visual representation learning.

\bibliography{reference}

\begin{thebibliography}{56}
\providecommand{\natexlab}[1]{#1}
\providecommand{\url}[1]{\texttt{#1}}
\expandafter\ifx\csname urlstyle\endcsname\relax
  \providecommand{\doi}[1]{doi: #1}\else
  \providecommand{\doi}{doi: \begingroup \urlstyle{rm}\Url}\fi

\bibitem[Baron et~al.(2023)Baron, Zimerman, and Wolf]{baron20232}
Ethan Baron, Itamar Zimerman, and Lior Wolf.
\newblock A 2-dimensional state space layer for spatial inductive bias.
\newblock In \emph{The Twelfth International Conference on Learning Representations}, 2023.

\bibitem[Cai \& Vasconcelos(2018)Cai and Vasconcelos]{cai2018cascade}
Zhaowei Cai and Nuno Vasconcelos.
\newblock Cascade r-cnn: Delving into high quality object detection.
\newblock In \emph{Proceedings of the IEEE conference on computer vision and pattern recognition}, pp.\  6154--6162, 2018.

\bibitem[Chen et~al.(2024)Chen, Huang, Xu, Pei, Chen, Li, Wang, Li, Lu, and Wang]{chen2024video}
Guo Chen, Yifei Huang, Jilan Xu, Baoqi Pei, Zhe Chen, Zhiqi Li, Jiahao Wang, Kunchang Li, Tong Lu, and Limin Wang.
\newblock Video mamba suite: State space model as a versatile alternative for video understanding.
\newblock \emph{arXiv preprint arXiv:2403.09626}, 2024.

\bibitem[Chen et~al.(2019)Chen, Wang, Pang, Cao, Xiong, Li, Sun, Feng, Liu, Xu, et~al.]{chen2019mmdetection}
Kai Chen, Jiaqi Wang, Jiangmiao Pang, Yuhang Cao, Yu~Xiong, Xiaoxiao Li, Shuyang Sun, Wansen Feng, Ziwei Liu, Jiarui Xu, et~al.
\newblock Mmdetection: Open mmlab detection toolbox and benchmark.
\newblock \emph{arXiv preprint arXiv:1906.07155}, 2019.

\bibitem[Chen et~al.(2020)Chen, Dai, Liu, Chen, Yuan, and Liu]{chen2020dynamic}
Yinpeng Chen, Xiyang Dai, Mengchen Liu, Dongdong Chen, Lu~Yuan, and Zicheng Liu.
\newblock Dynamic convolution: Attention over convolution kernels.
\newblock In \emph{Proceedings of the IEEE/CVF conference on computer vision and pattern recognition}, pp.\  11030--11039, 2020.

\bibitem[Contributors(2020)]{contributors2020mmsegmentation}
MMSegmentation Contributors.
\newblock Mmsegmentation: Openmmlab semantic segmentation toolbox and benchmark, 2020.

\bibitem[Dai et~al.(2017)Dai, Qi, Xiong, Li, Zhang, Hu, and Wei]{dai2017deformable}
Jifeng Dai, Haozhi Qi, Yuwen Xiong, Yi~Li, Guodong Zhang, Han Hu, and Yichen Wei.
\newblock Deformable convolutional networks.
\newblock In \emph{Proceedings of the IEEE international conference on computer vision}, pp.\  764--773, 2017.

\bibitem[Dao \& Gu(2024)Dao and Gu]{dao2024transformers}
Tri Dao and Albert Gu.
\newblock Transformers are ssms: Generalized models and efficient algorithms through structured state space duality.
\newblock \emph{arXiv preprint arXiv:2405.21060}, 2024.

\bibitem[Deng et~al.(2009)Deng, Dong, Socher, Li, Li, and Fei-Fei]{deng2009imagenet}
Jia Deng, Wei Dong, Richard Socher, Li-Jia Li, Kai Li, and Li~Fei-Fei.
\newblock Imagenet: A large-scale hierarchical image database.
\newblock In \emph{2009 IEEE conference on computer vision and pattern recognition}, pp.\  248--255. Ieee, 2009.

\bibitem[Ding et~al.(2022)Ding, Zhang, Han, and Ding]{ding2022scaling}
Xiaohan Ding, Xiangyu Zhang, Jungong Han, and Guiguang Ding.
\newblock Scaling up your kernels to 31x31: Revisiting large kernel design in cnns.
\newblock In \emph{Proceedings of the IEEE/CVF conference on computer vision and pattern recognition}, pp.\  11963--11975, 2022.

\bibitem[Dosovitskiy et~al.(2020)Dosovitskiy, Beyer, Kolesnikov, Weissenborn, Zhai, Unterthiner, Dehghani, Minderer, Heigold, Gelly, et~al.]{dosovitskiy2020vit}
Alexey Dosovitskiy, Lucas Beyer, Alexander Kolesnikov, Dirk Weissenborn, Xiaohua Zhai, Thomas Unterthiner, Mostafa Dehghani, Matthias Minderer, Georg Heigold, Sylvain Gelly, et~al.
\newblock An image is worth 16x16 words: Transformers for image recognition at scale.
\newblock \emph{arXiv preprint arXiv:2010.11929}, 2020.

\bibitem[Du et~al.(2022)Du, Zhou, Feng, Tan, and Zhou]{du2022sharpness}
Jiawei Du, Daquan Zhou, Jiashi Feng, Vincent Tan, and Joey~Tianyi Zhou.
\newblock Sharpness-aware training for free.
\newblock \emph{Advances in Neural Information Processing Systems}, 35:\penalty0 23439--23451, 2022.

\bibitem[Friston et~al.(2003)Friston, Harrison, and Penny]{friston2003dynamic}
Karl~J Friston, Lee Harrison, and Will Penny.
\newblock Dynamic causal modelling.
\newblock \emph{Neuroimage}, 19\penalty0 (4):\penalty0 1273--1302, 2003.

\bibitem[Fu et~al.(2022)Fu, Dao, Saab, Thomas, Rudra, and R{\'e}]{fu2022hungry}
Daniel~Y Fu, Tri Dao, Khaled~K Saab, Armin~W Thomas, Atri Rudra, and Christopher R{\'e}.
\newblock Hungry hungry hippos: Towards language modeling with state space models.
\newblock \emph{arXiv preprint arXiv:2212.14052}, 2022.

\bibitem[Gu \& Dao(2023)Gu and Dao]{gu2023mamba}
Albert Gu and Tri Dao.
\newblock Mamba: Linear-time sequence modeling with selective state spaces.
\newblock \emph{arXiv preprint arXiv:2312.00752}, 2023.

\bibitem[Gu et~al.(2020)Gu, Dao, Ermon, Rudra, and R{\'e}]{gu2020hippo}
Albert Gu, Tri Dao, Stefano Ermon, Atri Rudra, and Christopher R{\'e}.
\newblock Hippo: Recurrent memory with optimal polynomial projections.
\newblock \emph{Advances in neural information processing systems}, 33:\penalty0 1474--1487, 2020.

\bibitem[Gu et~al.(2021{\natexlab{a}})Gu, Goel, and R{\'e}]{gu2021efficiently}
Albert Gu, Karan Goel, and Christopher R{\'e}.
\newblock Efficiently modeling long sequences with structured state spaces.
\newblock \emph{arXiv preprint arXiv:2111.00396}, 2021{\natexlab{a}}.

\bibitem[Gu et~al.(2021{\natexlab{b}})Gu, Johnson, Goel, Saab, Dao, Rudra, and R{\'e}]{gu2021combining}
Albert Gu, Isys Johnson, Karan Goel, Khaled Saab, Tri Dao, Atri Rudra, and Christopher R{\'e}.
\newblock Combining recurrent, convolutional, and continuous-time models with linear state space layers.
\newblock \emph{Advances in neural information processing systems}, 34:\penalty0 572--585, 2021{\natexlab{b}}.

\bibitem[Guo et~al.(2024)Guo, Li, Dai, Ouyang, Ren, and Xia]{guo2024mambair}
Hang Guo, Jinmin Li, Tao Dai, Zhihao Ouyang, Xudong Ren, and Shu-Tao Xia.
\newblock Mambair: A simple baseline for image restoration with state-space model.
\newblock \emph{arXiv preprint arXiv:2402.15648}, 2024.

\bibitem[Guo et~al.(2022)Guo, Han, Wu, Tang, Chen, Wang, and Xu]{guo2022cmt}
Jianyuan Guo, Kai Han, Han Wu, Yehui Tang, Xinghao Chen, Yunhe Wang, and Chang Xu.
\newblock Cmt: Convolutional neural networks meet vision transformers.
\newblock In \emph{Proceedings of the IEEE/CVF conference on computer vision and pattern recognition}, pp.\  12175--12185, 2022.

\bibitem[Gupta et~al.(2022)Gupta, Gu, and Berant]{gupta2022diagonal}
Ankit Gupta, Albert Gu, and Jonathan Berant.
\newblock Diagonal state spaces are as effective as structured state spaces.
\newblock \emph{Advances in Neural Information Processing Systems}, 35:\penalty0 22982--22994, 2022.

\bibitem[Hafner et~al.(2019)Hafner, Lillicrap, Ba, and Norouzi]{hafner2019dream}
Danijar Hafner, Timothy Lillicrap, Jimmy Ba, and Mohammad Norouzi.
\newblock Dream to control: Learning behaviors by latent imagination.
\newblock \emph{arXiv preprint arXiv:1912.01603}, 2019.

\bibitem[Han et~al.(2024)Han, Wang, Xia, Han, Pu, Ge, Song, Song, Zheng, and Huang]{han2024demystify}
Dongchen Han, Ziyi Wang, Zhuofan Xia, Yizeng Han, Yifan Pu, Chunjiang Ge, Jun Song, Shiji Song, Bo~Zheng, and Gao Huang.
\newblock Demystify mamba in vision: A linear attention perspective.
\newblock \emph{arXiv preprint arXiv:2405.16605}, 2024.

\bibitem[Hassani et~al.(2023)Hassani, Walton, Li, Li, and Shi]{hassani2023neighborhood}
Ali Hassani, Steven Walton, Jiachen Li, Shen Li, and Humphrey Shi.
\newblock Neighborhood attention transformer.
\newblock In \emph{Proceedings of the IEEE/CVF Conference on Computer Vision and Pattern Recognition}, pp.\  6185--6194, 2023.

\bibitem[He et~al.(2024{\natexlab{a}})He, Bai, Zhang, He, Chen, Gan, Wang, Li, Tian, and Xie]{he2024mambaad}
Haoyang He, Yuhu Bai, Jiangning Zhang, Qingdong He, Hongxu Chen, Zhenye Gan, Chengjie Wang, Xiangtai Li, Guanzhong Tian, and Lei Xie.
\newblock Mambaad: Exploring state space models for multi-class unsupervised anomaly detection.
\newblock \emph{arXiv preprint arXiv:2404.06564}, 2024{\natexlab{a}}.

\bibitem[He et~al.(2017)He, Gkioxari, Doll{\'a}r, and Girshick]{he2017mask}
Kaiming He, Georgia Gkioxari, Piotr Doll{\'a}r, and Ross Girshick.
\newblock Mask r-cnn.
\newblock In \emph{Proceedings of the IEEE international conference on computer vision}, pp.\  2961--2969, 2017.

\bibitem[He et~al.(2024{\natexlab{b}})He, Han, Tang, Wang, Yang, Guo, and Wang]{he2024densemamba}
Wei He, Kai Han, Yehui Tang, Chengcheng Wang, Yujie Yang, Tianyu Guo, and Yunhe Wang.
\newblock Densemamba: State space models with dense hidden connection for efficient large language models.
\newblock \emph{arXiv preprint arXiv:2403.00818}, 2024{\natexlab{b}}.

\bibitem[Hu et~al.(2024)Hu, Baumann, Gui, Grebenkova, Ma, Fischer, and Ommer]{hu2024zigma}
Vincent~Tao Hu, Stefan~Andreas Baumann, Ming Gui, Olga Grebenkova, Pingchuan Ma, Johannes Fischer, and Bjorn Ommer.
\newblock Zigma: Zigzag mamba diffusion model.
\newblock \emph{arXiv preprint arXiv:2403.13802}, 2024.

\bibitem[Huang et~al.(2024)Huang, Pei, You, Wang, Qian, and Xu]{huang2024localmamba}
Tao Huang, Xiaohuan Pei, Shan You, Fei Wang, Chen Qian, and Chang Xu.
\newblock Localmamba: Visual state space model with windowed selective scan.
\newblock \emph{arXiv preprint arXiv:2403.09338}, 2024.

\bibitem[Kalman(1960)]{kalman1960new}
Rudolph~Emil Kalman.
\newblock A new approach to linear filtering and prediction problems.
\newblock 1960.

\bibitem[Katharopoulos et~al.(2020)Katharopoulos, Vyas, Pappas, and Fleuret]{katharopoulos2020transformers}
Angelos Katharopoulos, Apoorv Vyas, Nikolaos Pappas, and Fran{\c{c}}ois Fleuret.
\newblock Transformers are rnns: Fast autoregressive transformers with linear attention.
\newblock In \emph{International conference on machine learning}, pp.\  5156--5165. PMLR, 2020.

\bibitem[Li et~al.(2024)Li, Li, Wang, He, Wang, Wang, and Qiao]{li2024videomamba}
Kunchang Li, Xinhao Li, Yi~Wang, Yinan He, Yali Wang, Limin Wang, and Yu~Qiao.
\newblock Videomamba: State space model for efficient video understanding.
\newblock \emph{arXiv preprint arXiv:2403.06977}, 2024.

\bibitem[Liao et~al.(2024)Liao, Zhu, Wang, Pan, Wang, and Ma]{liao2024lightm}
Weibin Liao, Yinghao Zhu, Xinyuan Wang, Cehngwei Pan, Yasha Wang, and Liantao Ma.
\newblock Lightm-unet: Mamba assists in lightweight unet for medical image segmentation.
\newblock \emph{arXiv preprint arXiv:2403.05246}, 2024.

\bibitem[Lin et~al.(2014)Lin, Maire, Belongie, Hays, Perona, Ramanan, Doll{\'a}r, and Zitnick]{lin2014microsoft}
Tsung-Yi Lin, Michael Maire, Serge Belongie, James Hays, Pietro Perona, Deva Ramanan, Piotr Doll{\'a}r, and C~Lawrence Zitnick.
\newblock Microsoft coco: Common objects in context.
\newblock In \emph{Computer Vision--ECCV 2014: 13th European Conference, Zurich, Switzerland, September 6-12, 2014, Proceedings, Part V 13}, pp.\  740--755. Springer, 2014.

\bibitem[Liu et~al.(2024)Liu, Tian, Zhao, Yu, Xie, Wang, Ye, and Liu]{liu2024vmamba}
Yue Liu, Yunjie Tian, Yuzhong Zhao, Hongtian Yu, Lingxi Xie, Yaowei Wang, Qixiang Ye, and Yunfan Liu.
\newblock Vmamba: Visual state space model.
\newblock \emph{arXiv preprint arXiv:2401.10166}, 2024.

\bibitem[Liu et~al.(2021)Liu, Lin, Cao, Hu, Wei, Zhang, Lin, and Guo]{liu2021swin}
Ze~Liu, Yutong Lin, Yue Cao, Han Hu, Yixuan Wei, Zheng Zhang, Stephen Lin, and Baining Guo.
\newblock Swin transformer: Hierarchical vision transformer using shifted windows.
\newblock In \emph{Proceedings of the IEEE/CVF international conference on computer vision}, pp.\  10012--10022, 2021.

\bibitem[Liu et~al.(2022)Liu, Mao, Wu, Feichtenhofer, Darrell, and Xie]{liu2022convnext}
Zhuang Liu, Hanzi Mao, Chao-Yuan Wu, Christoph Feichtenhofer, Trevor Darrell, and Saining Xie.
\newblock A convnet for the 2020s.
\newblock In \emph{Proceedings of the IEEE/CVF conference on computer vision and pattern recognition}, pp.\  11976--11986, 2022.

\bibitem[Ma et~al.(2024)Ma, Li, and Wang]{ma2024umamba}
Jun Ma, Feifei Li, and Bo~Wang.
\newblock U-mamba: Enhancing long-range dependency for biomedical image segmentation.
\newblock \emph{arXiv preprint arXiv:2401.04722}, 2024.

\bibitem[Mo \& Tian(2024)Mo and Tian]{mo2024scaling}
Shentong Mo and Yapeng Tian.
\newblock Scaling diffusion mamba with bidirectional ssms for efficient image and video generation.
\newblock \emph{arXiv preprint arXiv:2405.15881}, 2024.

\bibitem[Nguyen et~al.(2022)Nguyen, Goel, Gu, Downs, Shah, Dao, Baccus, and R{\'e}]{nguyen2022s4nd}
Eric Nguyen, Karan Goel, Albert Gu, Gordon Downs, Preey Shah, Tri Dao, Stephen Baccus, and Christopher R{\'e}.
\newblock S4nd: Modeling images and videos as multidimensional signals with state spaces.
\newblock \emph{Advances in neural information processing systems}, 35:\penalty0 2846--2861, 2022.

\bibitem[Qiao et~al.(2024)Qiao, Yu, Guo, Chen, Zhao, Sun, Wu, and Liu]{qiao2024vl}
Yanyuan Qiao, Zheng Yu, Longteng Guo, Sihan Chen, Zijia Zhao, Mingzhen Sun, Qi~Wu, and Jing Liu.
\newblock Vl-mamba: Exploring state space models for multimodal learning.
\newblock \emph{arXiv preprint arXiv:2403.13600}, 2024.

\bibitem[Radosavovic et~al.(2020)Radosavovic, Kosaraju, Girshick, He, and Doll{\'a}r]{radosavovic2020regnet}
Ilija Radosavovic, Raj~Prateek Kosaraju, Ross Girshick, Kaiming He, and Piotr Doll{\'a}r.
\newblock Designing network design spaces.
\newblock In \emph{Proceedings of the IEEE/CVF conference on computer vision and pattern recognition}, pp.\  10428--10436, 2020.

\bibitem[Shi et~al.(2024)Shi, Xia, Jin, Wang, Zhao, Xia, Xiao, and Yang]{shi2024vmambair}
Yuan Shi, Bin Xia, Xiaoyu Jin, Xing Wang, Tianyu Zhao, Xin Xia, Xuefeng Xiao, and Wenming Yang.
\newblock Vmambair: Visual state space model for image restoration.
\newblock \emph{arXiv preprint arXiv:2403.11423}, 2024.

\bibitem[Smith et~al.(2022)Smith, Warrington, and Linderman]{smith2022simplified}
Jimmy~TH Smith, Andrew Warrington, and Scott~W Linderman.
\newblock Simplified state space layers for sequence modeling.
\newblock \emph{arXiv preprint arXiv:2208.04933}, 2022.

\bibitem[Touvron et~al.(2021)Touvron, Cord, Douze, Massa, Sablayrolles, and J{\'e}gou]{touvron2021deit}
Hugo Touvron, Matthieu Cord, Matthijs Douze, Francisco Massa, Alexandre Sablayrolles, and Herv{\'e} J{\'e}gou.
\newblock Training data-efficient image transformers \& distillation through attention.
\newblock In \emph{International conference on machine learning}, pp.\  10347--10357. PMLR, 2021.

\bibitem[Wang et~al.(2022)Wang, Xie, Li, Fan, Song, Liang, Lu, Luo, and Shao]{wang2022pvt}
Wenhai Wang, Enze Xie, Xiang Li, Deng-Ping Fan, Kaitao Song, Ding Liang, Tong Lu, Ping Luo, and Ling Shao.
\newblock Pvt v2: Improved baselines with pyramid vision transformer.
\newblock \emph{Computational Visual Media}, 8\penalty0 (3):\penalty0 415--424, 2022.

\bibitem[Williams \& Lawrence(2007)Williams and Lawrence]{2007Linear}
Robert~L. Williams and Douglas~A. Lawrence.
\newblock Linear state-space control systems || observability.
\newblock 10.1002/9780470117873:\penalty0 149--184, 2007.

\bibitem[Xiao et~al.(2018)Xiao, Liu, Zhou, Jiang, and Sun]{xiao2018unified}
Tete Xiao, Yingcheng Liu, Bolei Zhou, Yuning Jiang, and Jian Sun.
\newblock Unified perceptual parsing for scene understanding.
\newblock In \emph{Proceedings of the European conference on computer vision (ECCV)}, pp.\  418--434, 2018.

\bibitem[Xiao et~al.(2024)Xiao, Song, Huang, Wang, Song, Ge, Li, and Shan]{xiao2024grootvl}
Yicheng Xiao, Lin Song, Shaoli Huang, Jiangshan Wang, Siyu Song, Yixiao Ge, Xiu Li, and Ying Shan.
\newblock Grootvl: Tree topology is all you need in state space model.
\newblock \emph{arXiv preprint arXiv:2406.02395}, 2024.

\bibitem[Yang et~al.(2024)Yang, Chen, Espinosa, Ericsson, Wang, Liu, and Crowley]{yang2024plainmamba}
Chenhongyi Yang, Zehui Chen, Miguel Espinosa, Linus Ericsson, Zhenyu Wang, Jiaming Liu, and Elliot~J Crowley.
\newblock Plainmamba: Improving non-hierarchical mamba in visual recognition.
\newblock \emph{arXiv preprint arXiv:2403.17695}, 2024.

\bibitem[Yao et~al.(2024)Yao, Hong, Li, and Chanussot]{yao2024spectralmamba}
Jing Yao, Danfeng Hong, Chenyu Li, and Jocelyn Chanussot.
\newblock Spectralmamba: Efficient mamba for hyperspectral image classification.
\newblock \emph{arXiv preprint arXiv:2404.08489}, 2024.

\bibitem[Yu(2015)]{yu2015multi}
F~Yu.
\newblock Multi-scale context aggregation by dilated convolutions.
\newblock \emph{arXiv preprint arXiv:1511.07122}, 2015.

\bibitem[Yu \& Wang(2024)Yu and Wang]{yu2024mambaout}
Weihao Yu and Xinchao Wang.
\newblock Mambaout: Do we really need mamba for vision?
\newblock \emph{arXiv preprint arXiv:2405.07992}, 2024.

\bibitem[Yue \& Li(2024)Yue and Li]{yue2024medmamba}
Yubiao Yue and Zhenzhang Li.
\newblock Medmamba: Vision mamba for medical image classification.
\newblock \emph{arXiv preprint arXiv:2403.03849}, 2024.

\bibitem[Zhou et~al.(2019)Zhou, Zhao, Puig, Xiao, Fidler, Barriuso, and Torralba]{zhou2019semantic}
Bolei Zhou, Hang Zhao, Xavier Puig, Tete Xiao, Sanja Fidler, Adela Barriuso, and Antonio Torralba.
\newblock Semantic understanding of scenes through the ade20k dataset.
\newblock \emph{International Journal of Computer Vision}, 127:\penalty0 302--321, 2019.

\bibitem[Zhu et~al.(2024)Zhu, Liao, Zhang, Wang, Liu, and Wang]{zhu2024vision}
Lianghui Zhu, Bencheng Liao, Qian Zhang, Xinlong Wang, Wenyu Liu, and Xinggang Wang.
\newblock Vision mamba: Efficient visual representation learning with bidirectional state space model.
\newblock \emph{arXiv preprint arXiv:2401.09417}, 2024.

\end{thebibliography}
\bibliographystyle{iclr2025_conference}

\newpage
\begin{table*}[!t]
\centering
\Large
Appendix to ``Spatial-Mamba: Effective Visual State Space\\ Models via Structure-aware State Fusion"

\end{table*}
In this appendix, we provide the following materials:
\begin{itemize}[leftmargin=1em]
\item[\ref{sec:appendix:morevisual}] More visual results for SASF (referring to Sec. 4.1 in the main paper);
\item[\ref{sec:appendix:detailed}] Architecture details of Spatial-Mamba and experimental settings (referring to Sec. 4.2 and Sec. 5.1 in the main paper); 
\item[\ref{sec:appendix:derivation}] Derivations of the SSM formulas of Mamba and Spatial-Mamba (referring to Sec. 4.3 in the main paper);
\item[\ref{sec:appendix:cascademask}] Results of Cascade Mask R-CNN detector head (referring to Sec. 5.2 in the main paper);
\item[\ref{sec:appendix:visual}] 
Visual results and comparisons for detection and segmentation tasks (referring to Sec. 5.2 and Sec. 5.3 in the main paper);
\item[\ref{sec:appendix:extended ab study}] 
Further discussions about SASF (referring to Sec. 5.4 in the main paper);
\item[\ref{sec:appendix:erf}] Comparisons of effective receptive field (referring to Sec. 5.4 in the main paper).

\end{itemize}
\appendix

\section{More Visual Results}
\label{sec:appendix:morevisual}
\begin{figure}[h]
    \centering
    \begin{subfigure}{0.16\textwidth}
        \centering
        \includegraphics[height=1.8cm]{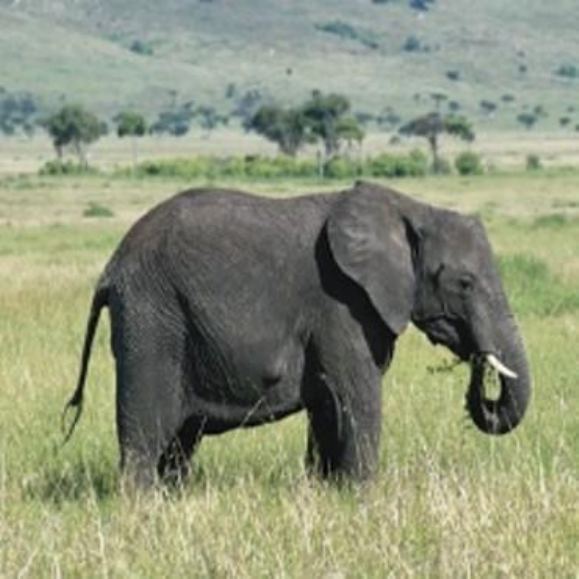}
    \end{subfigure}
    \begin{subfigure}{0.16\textwidth}
        \centering
        \includegraphics[height=1.8cm]{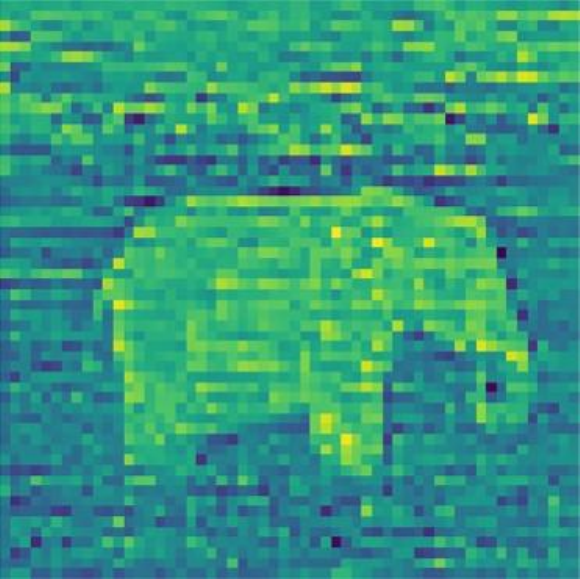}
    \end{subfigure}
    \begin{subfigure}{0.16\textwidth}
        \centering
        \includegraphics[height=1.8cm]{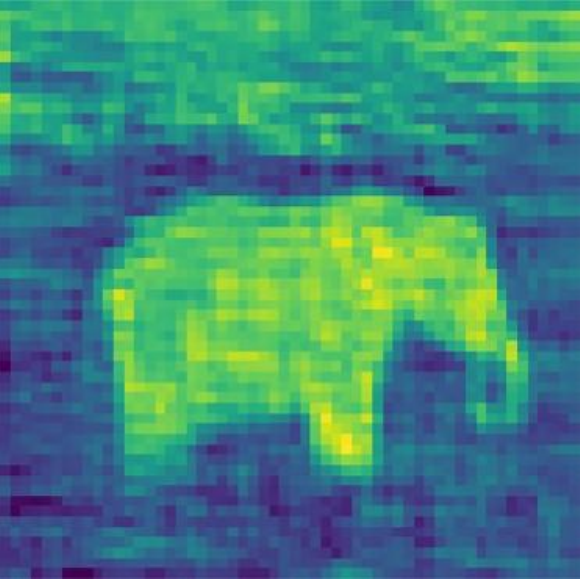}
    \end{subfigure}
    \begin{subfigure}{0.16\textwidth}
        \centering
        \includegraphics[height=1.8cm]{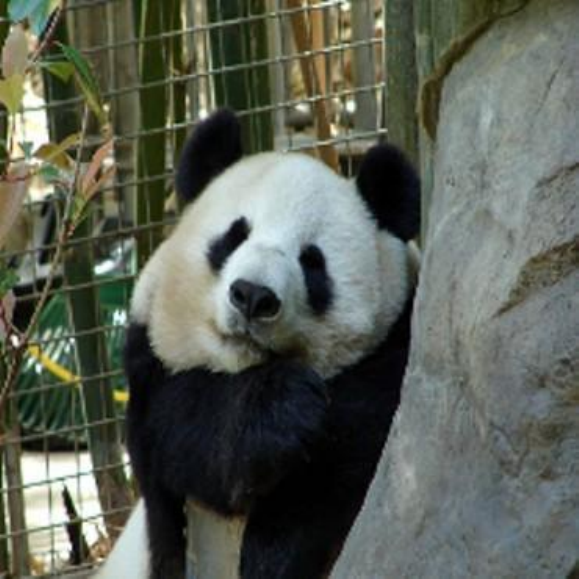}
    \end{subfigure}
    \begin{subfigure}{0.16\textwidth}
        \centering
        \includegraphics[height=1.8cm]{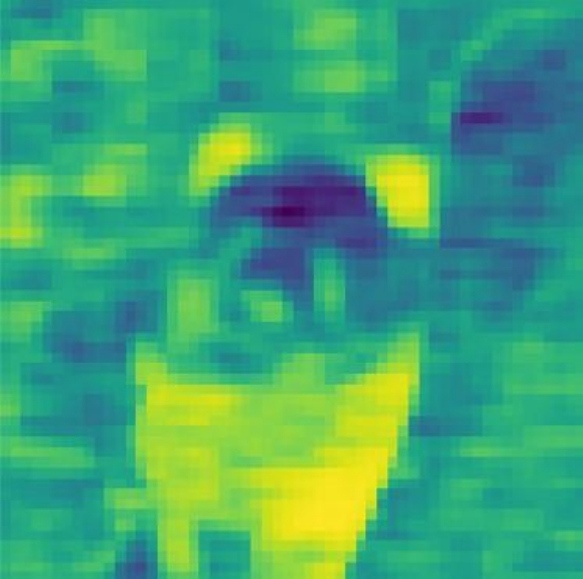}
    \end{subfigure}
    \begin{subfigure}{0.16\textwidth}
        \centering
        \includegraphics[height=1.8cm]{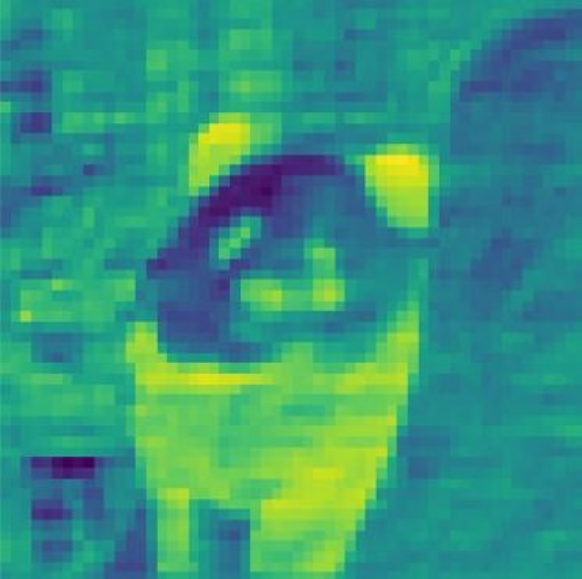}
    \end{subfigure}

    \begin{subfigure}{0.16\textwidth}
        \centering
        \includegraphics[height=1.8cm]{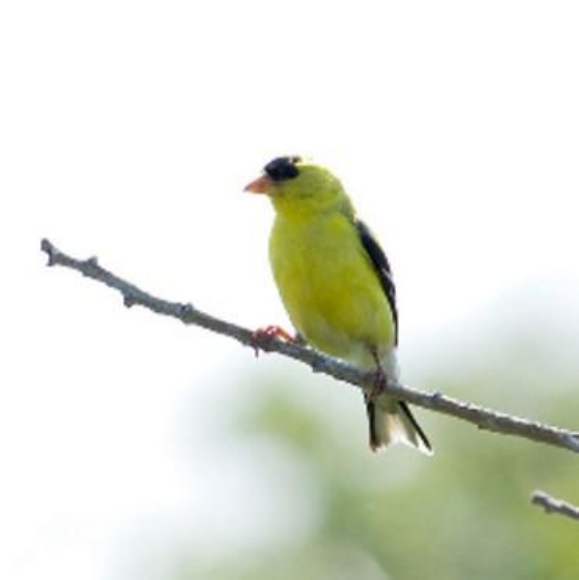}
    \end{subfigure}
    \begin{subfigure}{0.16\textwidth}
        \centering
        \includegraphics[height=1.8cm]{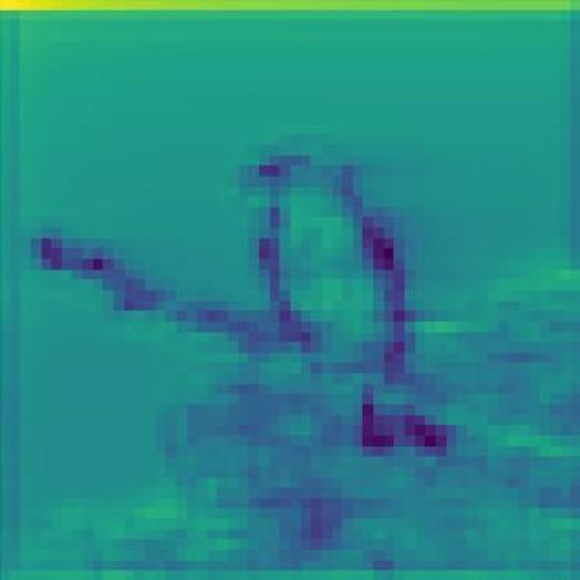}
    \end{subfigure}
    \begin{subfigure}{0.16\textwidth}
        \centering
        \includegraphics[height=1.8cm]{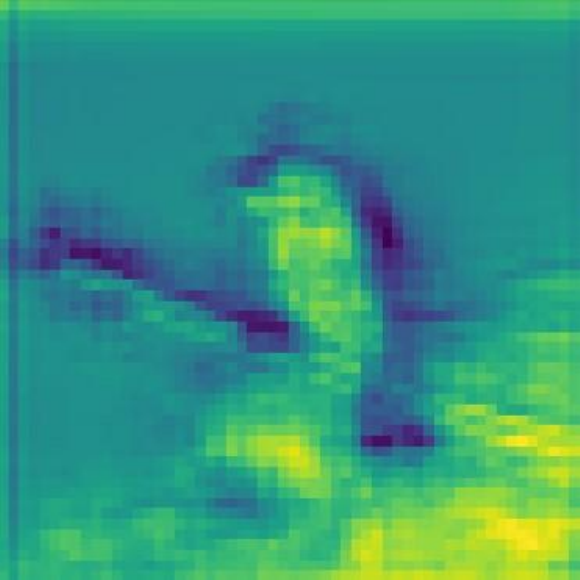}
    \end{subfigure}
    \begin{subfigure}{0.16\textwidth}
        \centering
        \includegraphics[height=1.8cm]{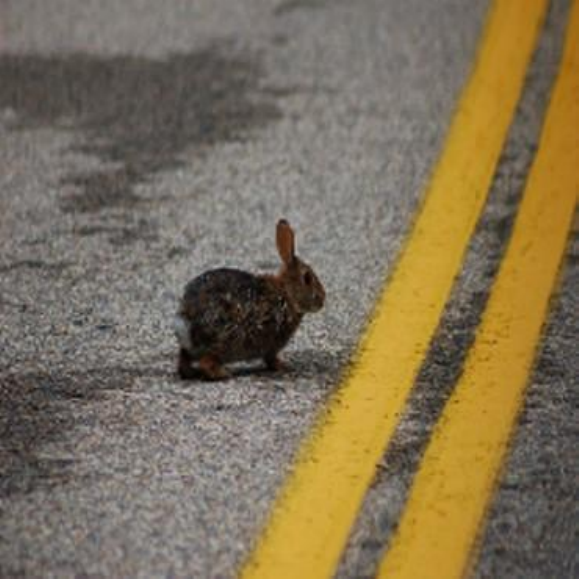}
    \end{subfigure}
    \begin{subfigure}{0.16\textwidth}
        \centering
        \includegraphics[height=1.8cm]{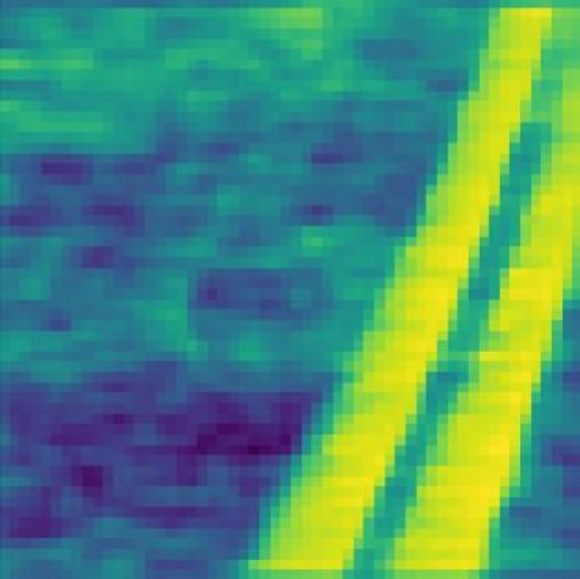}
    \end{subfigure}
    \begin{subfigure}{0.16\textwidth}
        \centering
        \includegraphics[height=1.8cm]{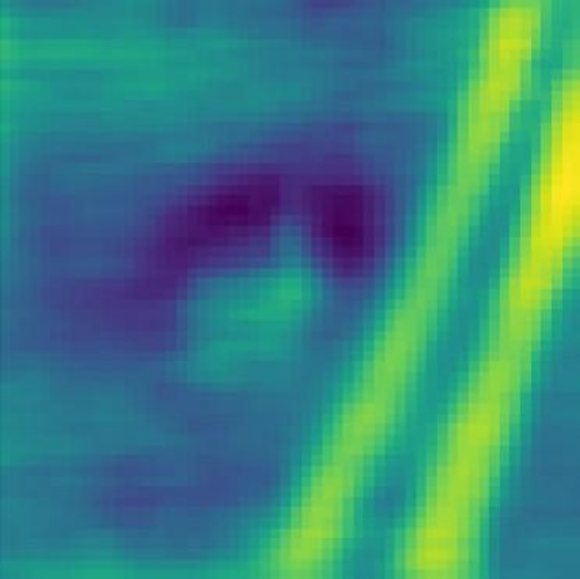}
    \end{subfigure}

    \begin{subfigure}{0.16\textwidth}
        \centering
        \includegraphics[height=1.8cm]{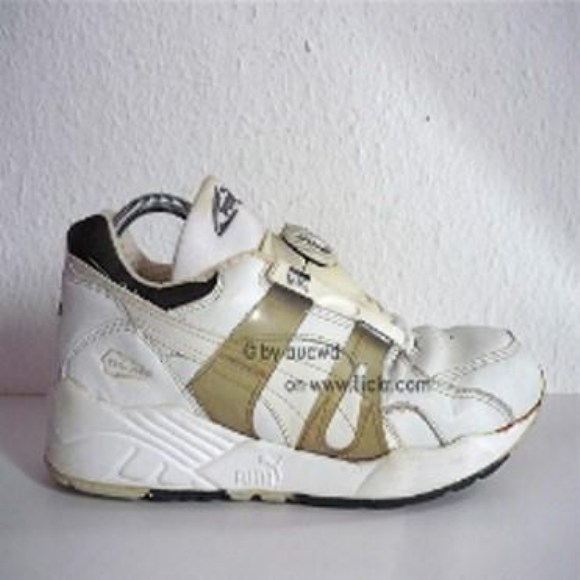}
        \caption{Input}
    \end{subfigure}
    \begin{subfigure}{0.16\textwidth}
        \centering
        \includegraphics[height=1.8cm]{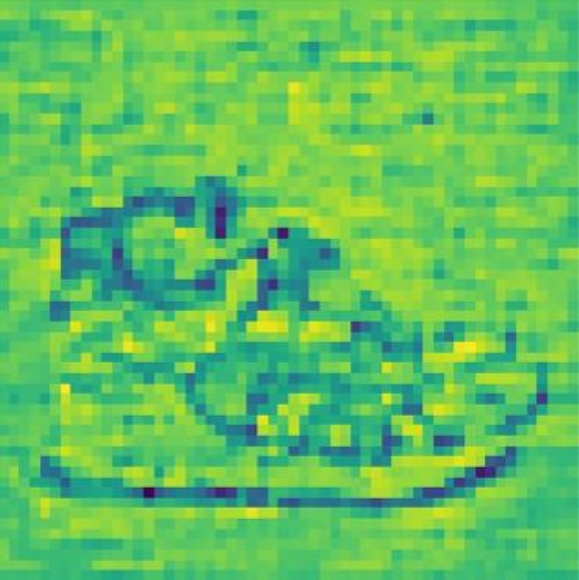}
        \caption{Original state}
    \end{subfigure}
    \begin{subfigure}{0.16\textwidth}
        \centering
        \includegraphics[height=1.8cm]{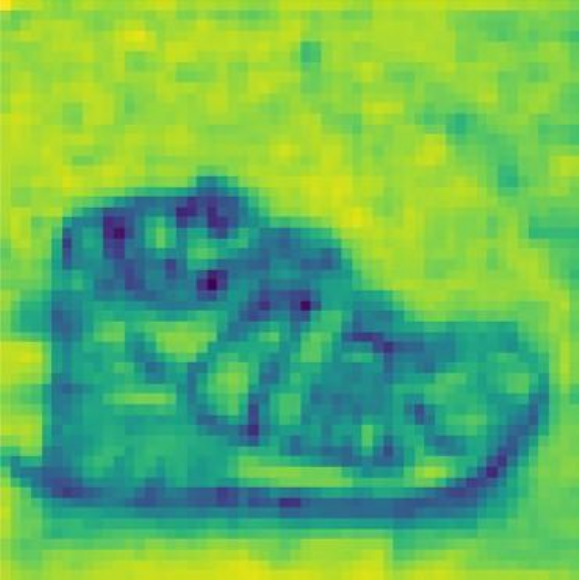}
        \caption{Fused state}
    \end{subfigure}
    \begin{subfigure}{0.16\textwidth}
        \centering
        \includegraphics[height=1.8cm]{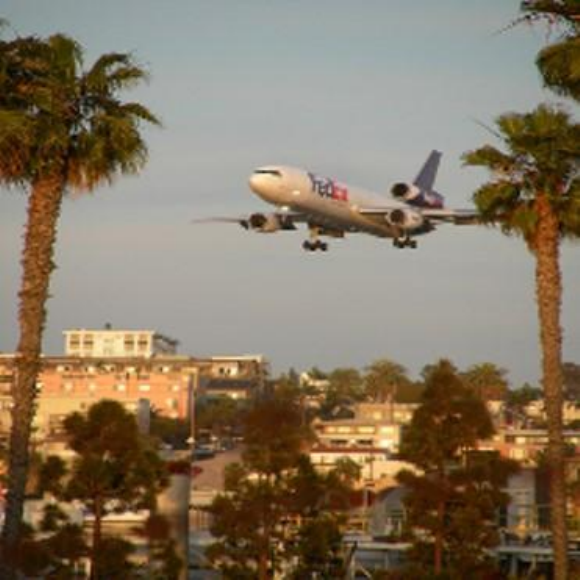}
        \caption{Input}
    \end{subfigure}
    \begin{subfigure}{0.16\textwidth}
        \centering
        \includegraphics[height=1.8cm]{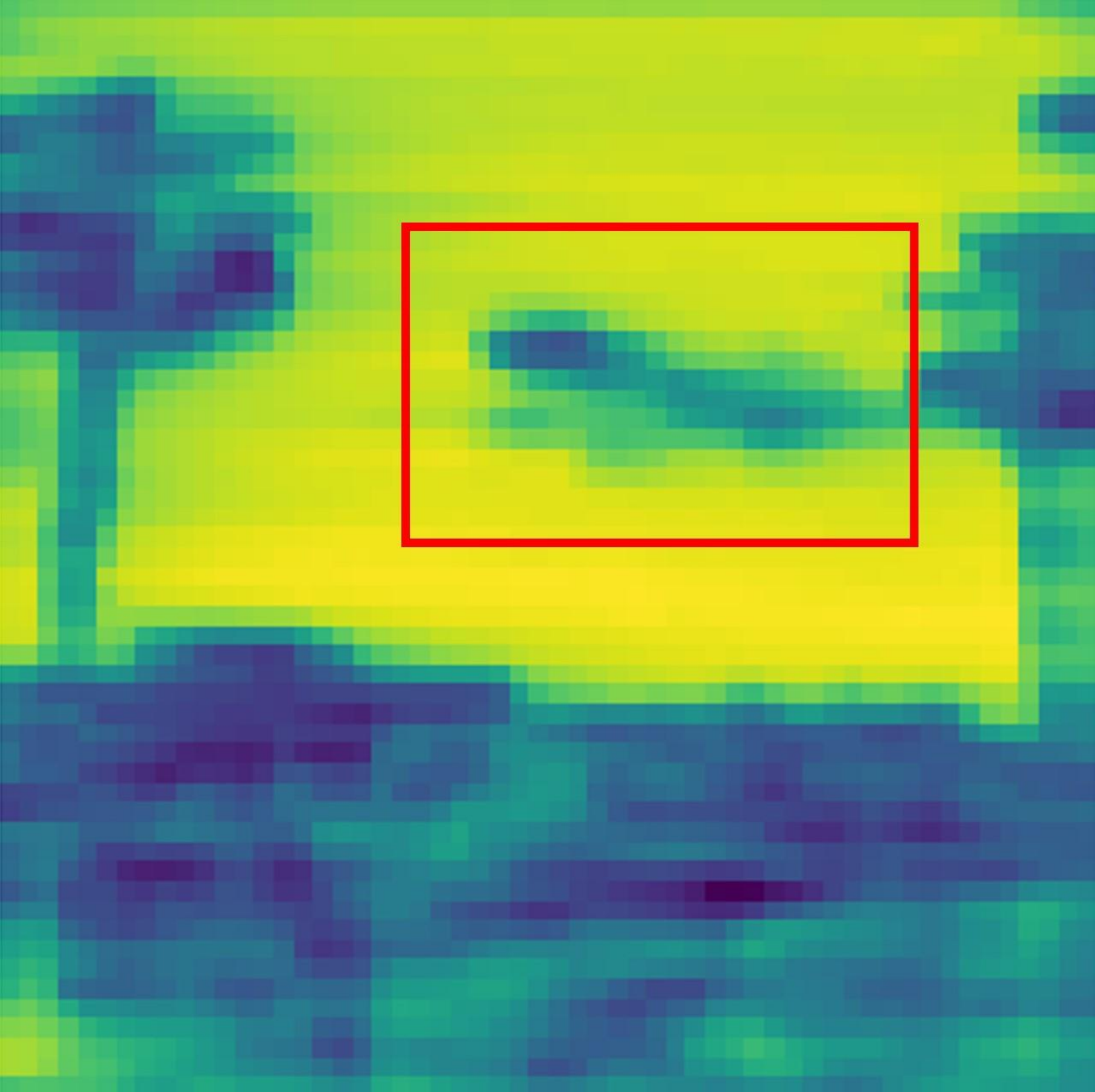}
        \caption{Original state}
    \end{subfigure}
    \begin{subfigure}{0.16\textwidth}
        \centering
        \includegraphics[height=1.8cm]{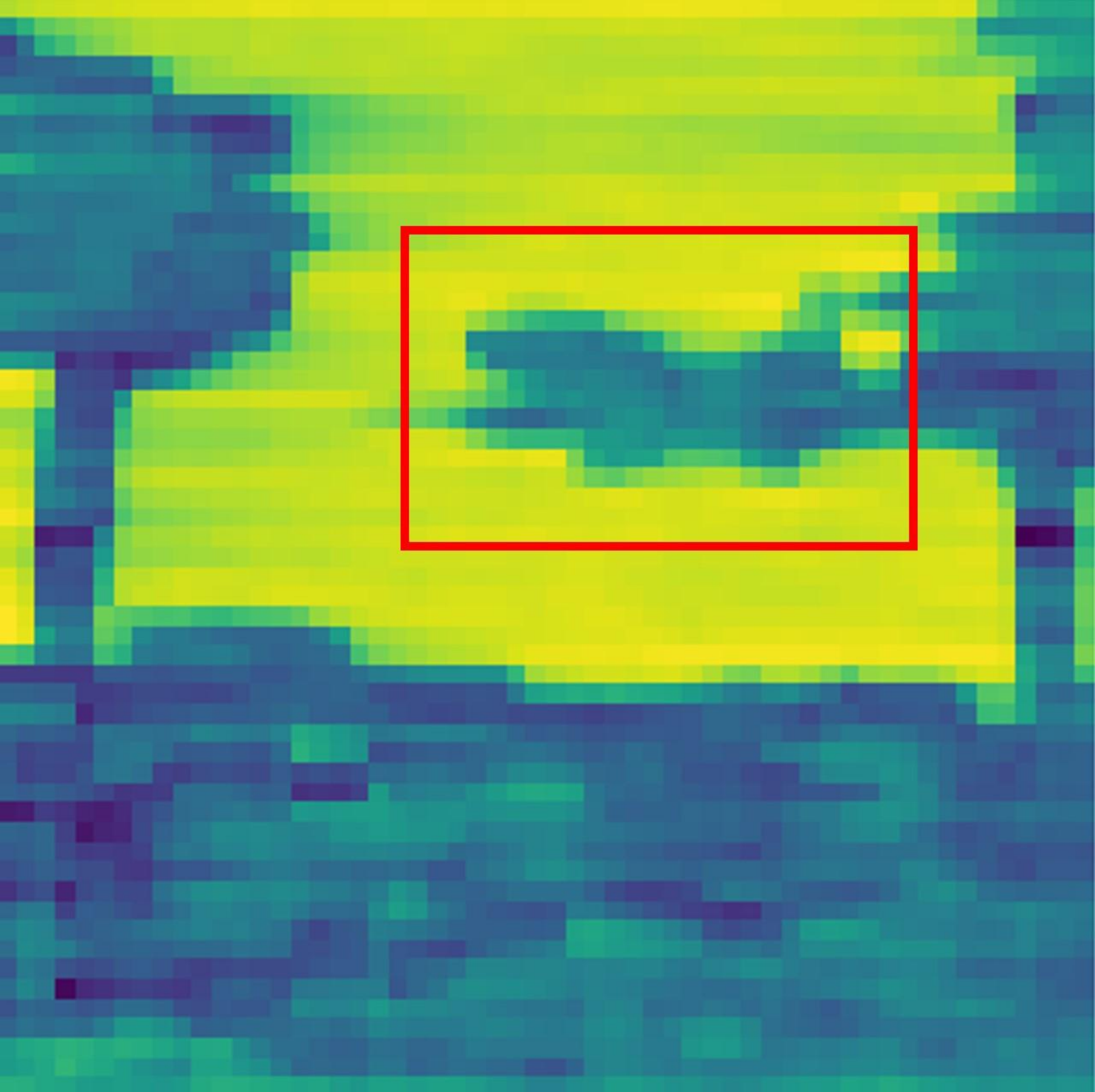}
        \caption{Fused state}
    \end{subfigure}
    
    \caption{More visualizations of state variables before and after applying the SASF equation.}
    \label{fig:morestatevs}
\end{figure}

More visual comparisons between the original and fused structure-aware state variables are shown in Fig.~\ref{fig:morestatevs}. Due to the spatial modeling capability of SASF, the fused state variables demonstrate more accurate context information and superior structural perception across different scenarios.

\section{Model Architectures and Experiment Settings in Classification}
\label{sec:appendix:detailed}
\begin{table}[!t]
    \centering
    \caption{Architectures of Spatial-Mamba models, where Linear refers to a linear layer, DWConv represents a depth-wise convolution layer, SASF module refers to Fig. 4 in Sec. 4.2 of the main paper, and FFN is a Feed-Forward Network.}
    \resizebox{\linewidth}{!}{
    \begin{tabular}{c|c|c|c|c}
    \toprule
    \textbf{Layer} &\textbf{Output size} & \textbf{Spatial-Mamba-T} & \textbf{Spatial-Mamba-S} & \textbf{Spatial-Mamba-B} \\ 
    \midrule
    
    Stem &  $56 \times 56$ & \multicolumn{3}{c}{  Conv $3\times 3$ stride 2, BN, GELU; Conv $3\times 3$ stride 1, BN; Conv $3\times 3$ stride 2, BN} \\ 
    \midrule
    \multirow{7}{*}{Stage1}& \multirow{7}{*}{ $28 \times 28$} & Spatial-Mamba Blocks & Spatial-Mamba Blocks & Spatial-Mamba Blocks \\
     & & $\left[\begin{array}{c}\text{Linear 64 $\rightarrow$ 128} \\ \text{DWConv 128} \\ \text{SASF 128} \\ \text{Linear 128 $\rightarrow$ 64} \\ \text{FFN 64}\end{array}\right] \times$ 2 & $\left[\begin{array}{c}\text{Linear 64 $\rightarrow$ 128} \\ \text{DWConv 128} \\ \text{SASF 128} \\ \text{Linear 128 $\rightarrow$ 64} \\ \text{FFN 64}\end{array}\right] \times$ 2 & $\left[\begin{array}{c}\text{Linear 96 $\rightarrow$ 192} \\ \text{DWConv 192} \\ \text{SASF 192} \\ \text{Linear 192 $\rightarrow$ 96} \\ \text{FFN 96}\end{array}\right] \times$ 2 \\ 
    \cline{3-5} 
     & & \multicolumn{3}{c}{ Down Sampling  Conv $3\times 3$ stride 2, LN} \\ 
    \midrule

    \multirow{7}{*}{Stage2}& \multirow{7}{*}{ $14 \times 14$} & Spatial-Mamba Blocks & Spatial-Mamba Blocks & Spatial-Mamba Blocks \\
     & & $\left[\begin{array}{c}\text{Linear 128 $\rightarrow$ 256} \\ \text{DWConv 256} \\ \text{SASF 256} \\ \text{Linear 256 $\rightarrow$ 128} \\ \text{FFN 128}\end{array}\right] \times$ 4 & $\left[\begin{array}{c}\text{Linear 128 $\rightarrow$ 256} \\ \text{DWConv 256} \\ \text{SASF 256} \\ \text{Linear 256 $\rightarrow$ 128} \\ \text{FFN 128}\end{array}\right] \times$ 4 & $\left[\begin{array}{c}\text{Linear 192 $\rightarrow$ 384} \\ \text{DWConv 384} \\ \text{SASF 384} \\ \text{Linear 384 $\rightarrow$ 192} \\ \text{FFN 192}\end{array}\right] \times$ 4 \\ 
    \cline{3-5} 
     & & \multicolumn{3}{c}{ Down Sampling  Conv $3\times 3$ stride 2, LN} \\ 
    \midrule

    \multirow{7}{*}{Stage3}& \multirow{7}{*}{ $7 \times 7$} & Spatial-Mamba Blocks & Spatial-Mamba Blocks & Spatial-Mamba Blocks \\
     & & $\left[\begin{array}{c}\text{Linear 256 $\rightarrow$ 512} \\ \text{DWConv 512} \\ \text{SASF 512} \\ \text{Linear 512 $\rightarrow$ 256} \\ \text{FFN 256}\end{array}\right] \times$ 8 & $\left[\begin{array}{c}\text{Linear 256 $\rightarrow$ 512} \\ \text{DWConv 512} \\ \text{SASF 512} \\ \text{Linear 512 $\rightarrow$ 256} \\ \text{FFN 256}\end{array}\right] \times$ 21 & $\left[\begin{array}{c}\text{Linear 384 $\rightarrow$ 768} \\ \text{DWConv 768} \\ \text{SASF 768} \\ \text{Linear 768 $\rightarrow$ 384} \\ \text{FFN 384}\end{array}\right] \times$ 21 \\ 
    \cline{3-5} 
     & & \multicolumn{3}{c}{ Down Sampling  Conv $3\times 3$ stride 2, LN} \\ 
    \midrule

    \multirow{6}{*}{Stage4}& \multirow{6}{*}{ $7 \times 7$} & Spatial-Mamba Blocks & Spatial-Mamba Blocks & Spatial-Mamba Blocks \\
     & & $\left[\begin{array}{c}\text{Linear 512 $\rightarrow$ 1024} \\ \text{DWConv 1024} \\ \text{SASF 1024} \\ \text{Linear 1024 $\rightarrow$ 512} \\ \text{FFN 512}\end{array}\right] \times$ 4 & $\left[\begin{array}{c}\text{Linear 512 $\rightarrow$ 1024} \\ \text{DWConv 1024} \\ \text{SASF 1024} \\ \text{Linear 1024 $\rightarrow$ 512} \\ \text{FFN 512}\end{array}\right] \times$ 5 & $\left[\begin{array}{c}\text{Linear 768 $\rightarrow$ 1536} \\ \text{DWConv 1536} \\ \text{SASF 1536} \\ \text{Linear 1536 $\rightarrow$ 768} \\ \text{FFN 768}\end{array}\right] \times$ 5 \\ 
    \midrule
    
    Head & $1 \times 1$ & \multicolumn{3}{c}{Average pool, Linear 1000, Softmax} \\ 
     
     \bottomrule
    \end{tabular}
    }
    \label{table:detailed arch}
\end{table}

\textbf{Network architecture.} The detailed architectures of Spatial-Mamba models are outlined in Tab.~\ref{table:detailed arch}. Following the common four-stage hierarchical framework \citep{liu2021swin,han2024demystify}, we construct the Spatial-Mamba models by stacking our proposed Spatial-Mamba blocks at each stage. Specifically, an input image with resolution of $224\times 224$ is firstly processed by a stem layer, which consists of Convolution (Conv), Batch Normalization (BN) and GELU activation function. The kernel size is $3\times 3$ with a stride of 2 at the first and last convolution layers, and a stride of 1 for other layers. Each stage contains multiple Spatial-Mamba blocks, followed by a down-sampling layer except for the last block. The down-sampling layer consists of a $3\times 3$ convolution with a stride of 2 and a Layer Normalization (LN) layer. Each block incorporates a structure-aware SSM layer and a Feed-Forward Network (FFN), both with residual connections. The structure-aware SSM contains a SASF branch with a 2D depth-wise convolution and a multiplicative gate branch with activation function, as illustrated in Sec. 4.2 of the main paper. The expand ratio of SSM is set to 2, doubling the number of channels. The SSM state dimension is set to 1 for better performance and efficiency. We modify the embedding dimension and number of blocks to build our Spatial-Mamba-T/S/B models.

\textbf{Settings for ImageNet-1K classification.} The Spatial-Mamba-T/S/B models are trained from scratch for 300 epochs using AdamW optimizer with betas set to (0.9, 0.999), momentum set to 0.9, and batch size set to 1024. The initial learning rate is set to 0.001 with a weight decay of 0.05. A cosine annealing learning rate schedule is adopted with a warm-up of 20 epochs. We adopt the common data augmentation strategies as in previous works \citep{liu2021swin,liu2024vmamba}.
Moreover, label smoothing (0.1), exponential moving average (EMA) and MESA \citep{du2022sharpness} are also applied. The drop path rate is set to 0.2 for Spatial-Mamba-T, 0.3 for Spatial-Mamba-S and 0.5 for Spatial-Mamba-B.

\textbf{Implementation details.} The Spatial-Mamba models employ a hardware-aware selective scan algorithm adapted from the original Mamba framework, with modifications to the CUDA kernels for decoupling state transition and observation equations. The SASF module of Spatial-Mamba is implemented by the general matrix multiplication (GEMM) with optimized CUDA kernels.

\section{Derivation of SSM Formulas}
\label{sec:appendix:derivation}

Based on the Mamba formulation provided in Sec. 3 of the main paper, the state transition equation in the recursive form can be rewritten as follows:
\begin{equation}\label{eq:appendix derivation rec}
\begin{aligned}
    x_t  &= \overline{\mA}_t x_{t-1} + \overline{\mB}_t u_t \\
     &= \overline{\mA}_t \left(\overline{\mA}_{t-1} x_{t-2} + \overline{\mB}_{t-1} u_{t-1}\right) + \overline{\mB}_t u_t\\
     &= \overline{\mA}_t \left(\overline{\mA}_{t-1} \left(\overline{\mA}_{t-2} x_{t-3} + \overline{\mB}_{t-2} u_{t-2}\right) + \overline{\mB}_{t-1} u_{t-1}\right) + \overline{\mB}_t u_t \\
     &=\overline{\mA}_t\overline{\mA}_{t-1}\overline{\mA}_{t-2}x_{t-3} + \overline{\mA}_t\overline{\mA}_{t-1}\overline{\mB}_{t-2}u_{t-2}+\overline{\mA}_t\overline{\mB}_{t-1}u_{t-1} + \overline{\mB}_t u_t \\
     &=\Pi_{i=1}^t\overline{\mA}_ix_0 + \Pi_{i=2}^t\overline{\mA}_i\overline{\mB}_1u_1 + \cdots + \Pi_{i=t-1}^t\overline{\mA}_i\overline{\mB}_{t-2}u_{t-2} + \Pi_{i=t}^t\overline{\mA}_i\overline{\mB}_{t-1}u_{t-1} + \overline{\mB}_t u_t\\
\end{aligned}
\end{equation}
By defining the initial state as zero, $x_0=0$, Eq. (\ref{eq:appendix derivation rec}) can be expressed as: 
\begin{equation}\label{eq:appendix derivation}
\begin{aligned}
    x_t  &=\sum_{s \leq t} \overline{\mA}_{s:t}^{\times}\ \overline{\mB}_s u_s, \quad \text{where } \overline{\mA}_{s:t}^{\times}:=\begin{cases} \Pi_{i=s+1}^t\overline{\mA}_{i}, & s<t \\1, & s=t\end{cases}.
\end{aligned}
\end{equation}

We omit the term $\mD_tu_t$ for simplicity. According to the observation equation, we can derive the final output $y_t=\sum_{s \leq t} \mC_t\ \overline{\mA}_{s:t}^{\times}\ \overline{\mB}_s u_s$. Suppose the input vector is $u=\left[u_1, \cdots, u_L\right]^T\in \mathbb{R}^L$ with length $L$, the corresponding output vector is $y\in\mathbb{R}^L$. Then the above calculation can be written in matrix multiplication form, \ie, $y=\mM u$, where $\mM$ is a structured lower triangular matrix and $\mM_{ij}=\mC_i \overline{\mA}_{j:i}^{\times}\overline{\mB}_j$.

Similarly, we can represent Spatial-Mamba in the same matrix transformation form. Based on the definition of Spatial-Mamba in Sec. 4.1 and Eq.~(\ref{eq:appendix derivation}), the SASF equation can be rewritten as $h_t=\sum_{k\in \Omega}\sum_{s \leq \rho_k(t)}\alpha_k \overline{\mA}_{s:\rho_k(t)}^{\times} \overline{\mB}_s u_s$. By multiplying $\mC_t$, we can derive the final output of Spatial-Mamba as $y_t  = \sum_{k\in \Omega}\sum_{s \leq \rho_k(t)}\alpha_k\mC_t\overline{\mA}_{s:\rho_k(t)}^{\times}\overline{\mB}_s u_s$. It can also be concisely represented as a matrix multiplication form $y=\mM u$, where $\mM$ is a structured adjacency matrix and $\mM_{ij}=\sum_{k}\mC_i \overline{\mA}_{j:\rho_k(i)}^{\times}\overline{\mB}_j$.

\section{Results of Cascade Mask R-CNN Detector Head}
\label{sec:appendix:cascademask}
Detailed results of object detection and instance segmentation on the COCO dataset with Cascade Mask R-CNN framework are reported in Tab.~\ref{table:cascade r-cnn}. We can see that Spatial-Mamba-T achieves a box mAP of 52.1 and a mask mAP of 44.9, surpassing Swin-T/NAT-T by 1.7/0.7 in box mAP and 1.2/0.4 in mask mAP with fewer parameters and FLOPs, respectively. Similarly, Spatial-Mamba-S demonstrates superior performance under the same configuration.

\begin{table}[!h]
    \centering
    \caption{Results of object detection and instance segmentation on the COCO dataset using Cascade Mask R-CNN \citep{cai2018cascade} under 3$\times$ schedule. FLOPs are calculated with input resolution of 1280 × 800.}
    \setlength{\tabcolsep}{1.7mm}
        \begin{tabular}{c|ccc|ccc|cc}
        \toprule
        Backbone & AP$^\text{b}\uparrow$ & AP$^\text{b}_{50}$$\uparrow$ & AP$^\text{b}_{75}$$\uparrow$ & AP$^\text{m}\uparrow$ & AP$^\text{m}_{50}$$\uparrow$ & AP$^\text{m}_{75}$$\uparrow$ & \#Param. & FLOPs \\
        \midrule
        Swin-T     & 50.4 & 69.2 & 54.7 & 43.7 & 66.6 & 47.3 & 86M & 745G \\
        ConvNeXt-T & 50.4 & 69.1 & 54.8 & 43.7 & 66.5 & 47.3 & 86M & 741G \\
        NAT-T      & 51.4 & 70.0 & 55.9 & 44.5 & 67.6 & 47.9 & 85M & 737G \\
        Spatial-Mamba-T   & \textbf{52.1} & \textbf{71.0} & \textbf{56.5} & \textbf{44.9} & \textbf{68.3} & \textbf{48.7} & 84M & 740G \\
        \midrule
        Swin-S     & 51.9 & 70.7 & 56.3 & 45.0 & 68.2 & 48.8 & 107M & 838G \\
        ConvNeXt-S & 51.9 & 70.8 & 56.5 & 45.0 & 68.4 & 49.1 & 108M & 827G \\
        NAT-S      & 52.0 & 70.4 & 56.3 & 44.9 & 68.1 & 48.6 & 108M & 809G \\
        Spatial-Mamba-S   & \textbf{53.3} & \textbf{71.9} & \textbf{57.9} & \textbf{45.8} & \textbf{69.4} & \textbf{49.7} & 101M & 794G \\
        \bottomrule 
        \end{tabular}
    \label{table:cascade r-cnn}
\end{table}

\section{Qualitative Results}
\label{sec:appendix:visual}
\begin{figure}[ht]
    \centering
    \begin{subfigure}{0.24\textwidth}
        \centering
        \includegraphics[height=7.0cm]{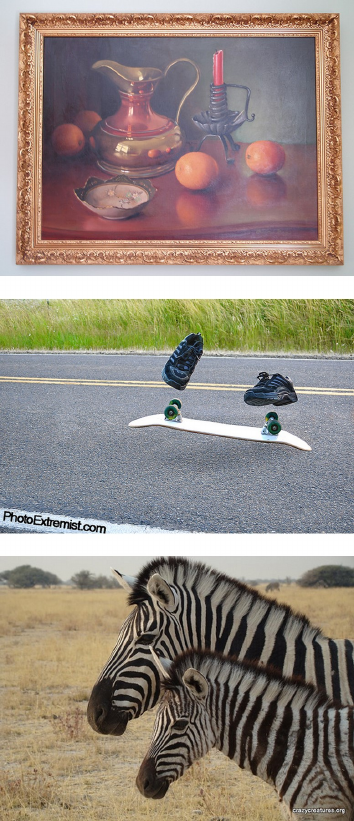}
        \caption{Input}
    \end{subfigure}
    \begin{subfigure}{0.24\textwidth}
        \centering
        \includegraphics[height=7.0cm]{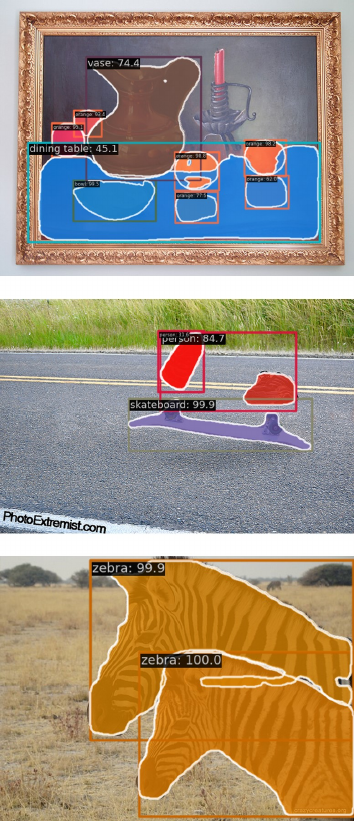}
        \caption{VMamba}
    \end{subfigure}
    \begin{subfigure}{0.24\textwidth}
        \centering
        \includegraphics[height=7.0cm]{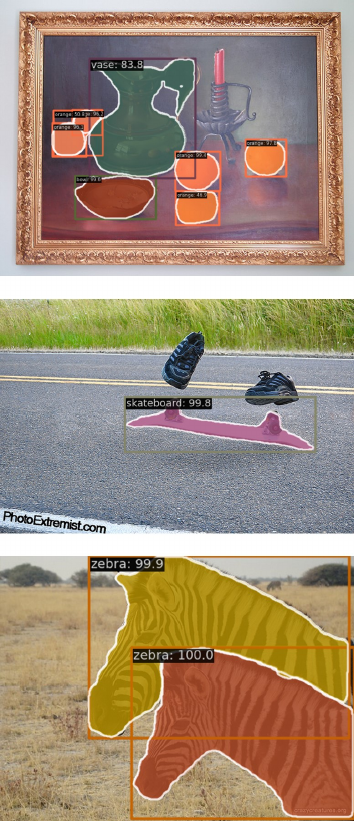}
        \caption{Spatial-Mamba}
    \end{subfigure}
    \begin{subfigure}{0.24\textwidth}
        \centering
        \includegraphics[height=7.0cm]{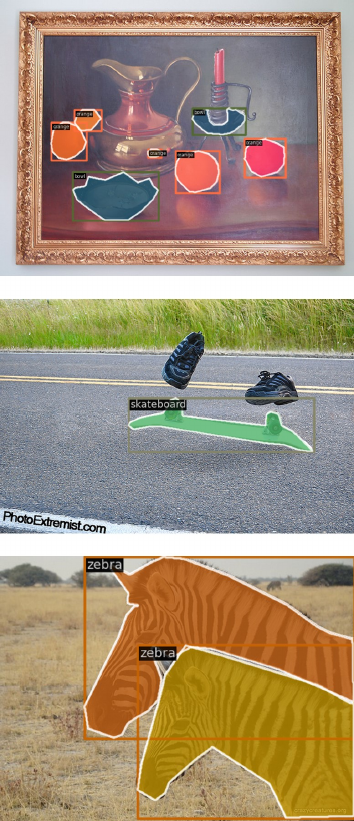}
        \caption{Ground Truth}
    \end{subfigure}
    \caption{Visualization examples of object detection and instance segmentation on COCO dataset with Mask R-CNN 1$\times$ schedule \citep{he2017mask} by VMamba and Spatial-Mamba.}
    \label{fig:detection}
\end{figure}

\begin{figure}[ht]
    \centering
    \begin{subfigure}{0.24\textwidth}
        \centering
        \includegraphics[height=7.0cm]{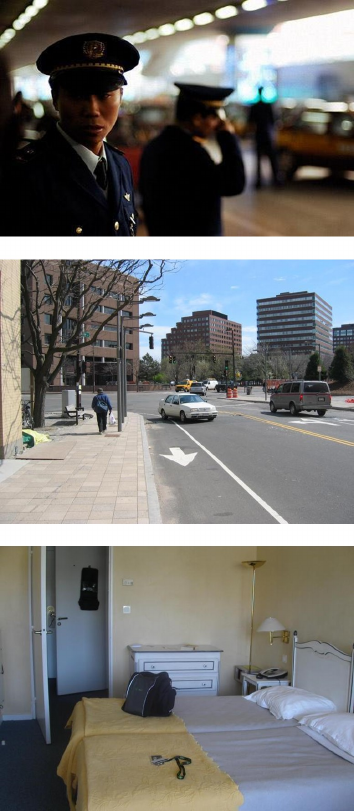}
        \caption{Input}
    \end{subfigure}
    \begin{subfigure}{0.24\textwidth}
        \centering
        \includegraphics[height=7.0cm]{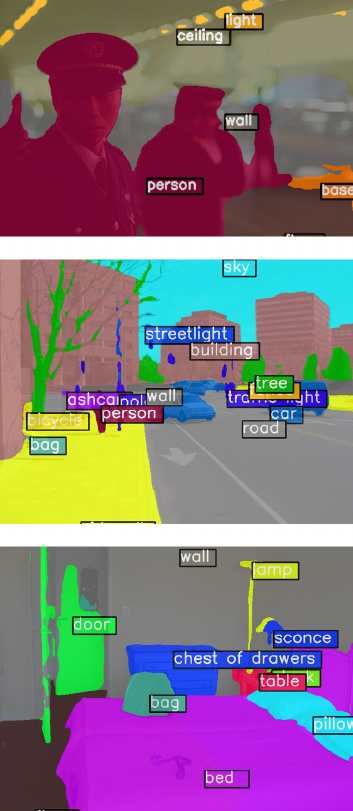}
        \caption{VMamba}
    \end{subfigure}
    \begin{subfigure}{0.24\textwidth}
        \centering
        \includegraphics[height=7.0cm]{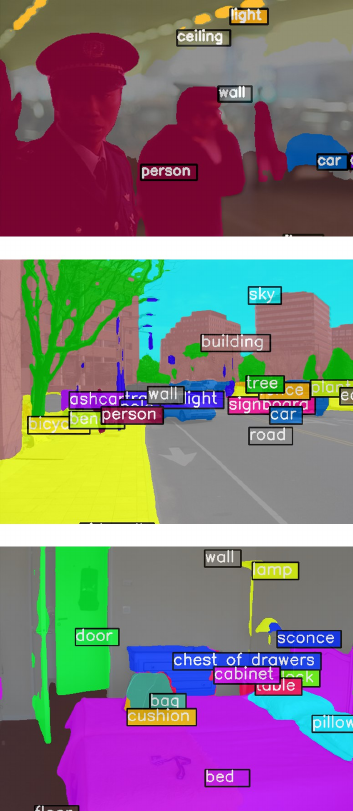}
        \caption{Spatial-Mamba}
    \end{subfigure}
    \begin{subfigure}{0.24\textwidth}
        \centering
        \includegraphics[height=7.0cm]{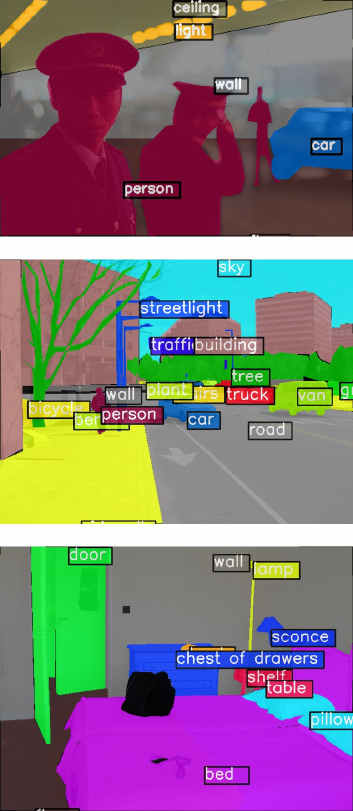}
        \caption{Ground Truth}
    \end{subfigure}
    \caption{Visualization examples of semantic segmentation on ADE20K dataset with single-scale inputs by VMamba and Spatial-Mamba.}
    \label{fig:segmentation}
\end{figure}

In this section, we present the visualization results of object detection and instance segmentation in Fig.~\ref{fig:detection}, and present the results of semantic segmentation in Fig.~\ref{fig:segmentation}. Compared with VMamba, our Spatial-Mamba demonstrates superior performance in both tasks, producing more accurate detection boxes segmentation masks, particularly in areas where local structural information is crucial. For example, in the second row of Fig.~\ref{fig:detection}, VMamba mistakenly identifies the shoes on a skateboard as a person, probably because it observes the shoes from four directions independently and resembles them as a human. Our method avoids this mistake by simultaneously perceiving the shoes and their surrounding context. Similarly, in the semantic segmentation task, as shown in the second and third rows of Fig.~\ref{fig:segmentation}, our approach achieves more precise structures of trees and doors. These results highlight the effectiveness of our proposed Spatial-Mamba in leveraging local structural information for better visual understanding.

\section{Further Discussions about SASF}
\label{sec:appendix:extended ab study}
\textbf{SASF extension.} To further explore and validate the spatial modeling capability of our proposed SASF, we adapt SASF into Vim and VMamba backbones, resulting in Vim+SASF (by replacing the middle class token with average pooling and setting state dimension to 1) and VMamba+SASF (directly integrated with SASF). Under the same training settings, the classification results on ImageNet-1K are presented in Tab.~\ref{table:sasf extention}. It can be seen that by integrating with SASF, the bi-directional Vim model is improved by 0.5\% in top-1 accuracy and the cross-scan VMamba is improved by 0.3\%. These findings verify that SASF can even enhance the performance of multi-directional models, which have already incorporated (yet in an inefficient manner) spatial modeling operations. 

\begin{table}[t]
\vspace{-6pt}
    \centering
    \caption{ Comparison of classification performance on ImageNet-1K by integrating with SASF, where `Throughput' is measured using an A100 GPU with an input resolution of $224\times 224$.}
    \setlength{\tabcolsep}{1.2mm}
        \begin{tabular}{l|cccc|c}
        \toprule
        Method & Im. size & \#Param.  (M) & FLOPs (G) & Throughput$\uparrow$ &  Top-1 acc.$\uparrow$ \\
        \midrule
        Vim-Ti  & $224^2$ & 6M & 1.4G  & -  & 71.7 \\
        Vim-Ti+SASF  & $224^2$ & 7M & 1.6G  & -  & \textbf{72.2}  \\
        VMamba-T & $224^2$ & 30M & 4.9G  & 1686  & 82.6 \\
        VMamba-T+SASF & $224^2$ & 31M  & 5.1G & 1126  & \textbf{82.9} \\
        \bottomrule
        \end{tabular}
        \label{table:sasf extention}
\vspace{-6pt}
\end{table}

\textbf{Fusion operators.} There are also different choices of the operators that can be used for fusing state variables in SASF. We opt for depth-wise convolution due to its simplicity and computational efficiency. However, operators like dynamic convolution \citep{chen2020dynamic}, deformable convolution \citep{dai2017deformable}, and attention mechanisms have also demonstrated superior performance in computer vision tasks. Therefore, we anticipate that they can be used as alternatives to the depth-wise convolution in our Spatial-Mamba. We simply train a Spatial-Mamba-T network with dynamic convolution and the results are shown Tab.~\ref{table:fusion op}. We can see that dynamic convolution yields a 0.2\% performance gain in accuracy but significantly reduces throughput from 1438 to 907. This result suggests the potential advantages of more flexible fusion operators for SASF, but also highlights the importance of considering computational cost.

\begin{table}[h]
    \centering
    \caption{ Comparison of classification performance by Spatial-Mamba-T with depth-wise conv. and dynamic conv. on ImageNet-1K, where `Throughput' is measured using an A100 GPU with an input resolution of $224\times 224$.} 
    
    \setlength{\tabcolsep}{1.2mm}
        \begin{tabular}{c|cccc|c}
        \toprule
        Method & Im. size & \#Param.  (M) & FLOPs (G) & Throughput$\uparrow$ &  Top-1 acc.$\uparrow$ \\
        \midrule
        Depth-wise Conv. & $224^2$ & 27M & 4.5G &  1438 & 83.5 \\
        Dynamic Conv. & $224^2$ & 33M & 4.8G &  907 & 83.7 \\
        \bottomrule
        \end{tabular}
    \label{table:fusion op}
\end{table}
\vspace{-6pt}

\section{Effective Receptive Field (ERF)}
\label{sec:appendix:erf}

\begin{figure}[!h]
    \centering
    \includegraphics[width=1\linewidth]{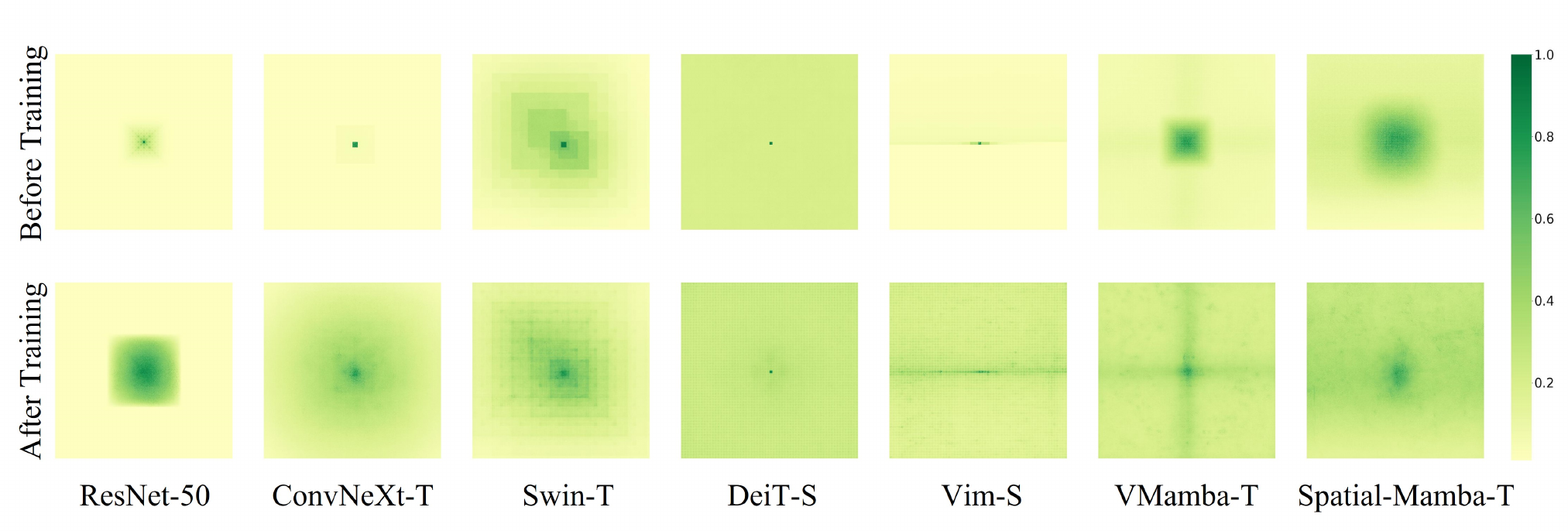}
    \caption{Comparison of Effective Receptive Field (ERF) among popular backbone networks.}
    \label{fig:ERF}
\end{figure}

We compare the Effective Receptive Field (ERF) \citep{ding2022scaling} of the center pixel on popular backbone networks before and after training, as shown in Fig.~\ref{fig:ERF}. 
The ERF values represent the contributions of every pixel on input space to the central pixel in the final output feature maps. 
To visualization, we randomly select 50 images from the ImageNet-1K validation set, resize them to a resolution of 1024$\times$1024, and then calculate the ERF values with the auto-grad mechanism. 
Before training, our Spatial-Mamba-T initially exhibits a larger receptive field than other methods except DeiT-S due to neighborhood connectivity in the state space. After training, our method, along with DeiT-S, Vim-S, and VMamba-T, all demonstrate a global ERF. In addition, both Vim-S and VMamba-T exhibit noticeable accumulation contributions along either horizontal or vertical directions, which can be attributed to their multi-directional fusion mechanisms. In contrast, our unidirectional Spatial-Mamba-T effectively eliminates this directional bias.

\end{document}